\pdfoutput=1

\newif\REVIEW

\ifdefined\REVIEW
    \documentclass[
        a4paper,
    ]{article}
    \usepackage{lineno, setspace}
    \modulolinenumbers[1]
\else
    \documentclass[
        a4paper,
        twocolumn
    ]{article}
\fi

\usepackage{IEK10} 
\usepackage{natbib}
\usepackage[section]{placeins}
\usepackage{comment}

\def\IEK10{
  Institute of Climate and Energy Systems,
  Energy Systems Engineering (ICE-1),
  Forschungszentrum J\"ulich GmbH,
  J\"ulich 52425,
  Germany
}
\def\RWTH{
  RWTH Aachen University,
  Aachen 52062,
  Germany
}
\def\JARA{
  JARA-ENERGY,
  J{\"u}lich 52425,
  Germany
}
\def\SVT{
  RWTH Aachen University,
  Process Systems Engineering (AVT.SVT),
  Aachen 52074,
  Germany
}
\def\FVT{
  RWTH Aachen University,
  Fluid Process Engineering (AVT.FVT),
  Aachen 52074,
  Germany
}
\def\IBG{
  Institute for Bio- and Geosciences (IBG-2),
  Forschungszentrum J\"ulich GmbH,
  J\"ulich 52425,
  Germany
}

\newcommand{\mytitle}{Estimating Dense-Packed Zone Height in Liquid-Liquid Separation: A Physics-Informed Neural Network Approach}

\newcommand{\affil}{
  \begin{itemize}[leftmargin=3mm, itemsep=0mm]
    \item[$^a$]\IEK10
    \item[$^b$]\RWTH
    \item[$^c$]\JARA
    \item[$^d$]\SVT
    \item[$^e$]\FVT
    \item[$^f$]\IBG
  \end{itemize}
}

\def\firstAuthor{Mehmet Velioglu}

\newcommand{\myauthor}{\firstAuthor$^{a,b}$,
Song Zhai$^{e}$,
Alexander Mitsos$^{c,a,d}$,
Adel Mhamdi$^{d}$,
Andreas Jupke$^{e,f}$,
Manuel Dahmen$^{a,*}$
}
\newcommand{\correspondingauthor}{M. Dahmen, \IEK10 \newline E-mail: m.dahmen@fz-juelich.de}
\author{\myauthor}

\usepackage[
  colorlinks,
  linkcolor=blue,
  citecolor=blue,
  urlcolor=blue,
  pdftitle={\mytitle},
  pdfauthor={\firstAuthor}
]{hyperref}
\usepackage{svg}
\usepackage[capitalise, nameinlink]{cleveref}
\crefname{table}{Tab.}{Tab.}

\begin{document}

\doublespacing

\thispagestyle{firststyle}

\begin{center}
    \begin{large}
      \textbf{\mytitle}
    \end{large} \\
    \myauthor
\end{center}

\vspace{0.5cm}

\begin{footnotesize}
    \affil
\end{footnotesize}

\vspace{0.5cm}
    
\begin{abstract} 
Separating liquid–liquid dispersions in gravity settlers is critical in chemical, pharmaceutical, and recycling processes. The dense-packed zone height is an important performance and safety indicator but it is often expensive and impractical to measure due to optical limitations. We propose a framework to estimate phase heights by combining a PINN model with readily available volume flow measurements, without requiring phase height measurements during deployment. To this end, a physics-informed neural network (PINN) is first pretrained on synthetic data and physics equations derived from a low-fidelity (approximate) mechanistic model to reduce the need for extensive experimental data. While the mechanistic model is used to generate synthetic training data, only volume balance equations are used in the PINN, as incorporating droplet coalescence and sedimentation submodels would be computationally prohibitive. The pretrained PINN is then fine-tuned with scarce experimental phase height and flow-rate data to capture the actual dynamics of the separator. We then deploy the differentiable PINN as a predictive model in an Extended Kalman Filter inspired state estimation framework, enabling the phase heights to be tracked and updated using flow-rate measurements only. We first test the two-stage trained PINN by forward simulation from a known initial state against the mechanistic model and a non-pretrained PINN. We then evaluate phase height estimation performance with the filter, comparing the two-stage trained PINN with a two-stage trained purely data-driven neural network. All model types are trained and evaluated using ensembles to account for model parameter uncertainty. In all evaluations, the two-stage trained PINN yields the most accurate phase-height estimates.
\end{abstract}

\vspace{0.5cm}

\noindent \textbf{Keywords}: \textit{Liquid-liquid separation, Physics-informed neural networks, State estimation, Extended Kalman Filter}

\vspace{0.5cm}

\section{Introduction}\label{sec:intro} 

The separation of liquid-liquid dispersions in a horizontal gravity separator is an essential step in many processes, such as solvent extraction in the chemical, pharmaceutical, and recycling industries, as well as grease and oil separation in the food industry and water treatment \citep{Frising.2006}. Liquid-liquid dispersions entering these separators split into their coherent phases under the influence of gravity-driven sedimentation and coalescence. Material systems characterized by a slower coalescence than sedimentation rate form a dispersion layer known as a dense-packed zone (DPZ) \citep{Henschke.2002}. Accumulation of the DPZ height above a critical level floods the separator, disrupting the separation process and leading to complications in subsequent unit operations due to poor separation efficiency \citep{Zhai.2025, Ye.2023b,Ye.2024}. Measuring the DPZ height and controlling it below a critical height is of utmost importance to prevent flooding, yet gauge glasses in industrial separators allow only limited capacity to measure the DPZ height \citep{Cusack2009}. 

Estimating the DPZ height by a suitable liquid-liquid gravity separator (LLS) model has the potential to detect flooding caused by DPZ accumulation. The steady-state LLS model of \citet{Henschke.1995} remains the state-of-the-art in describing the DPZ distribution along the LLS but dynamic models are limited and not validated \citep{Kamp.2017}. \citet{Backi2018AFirst-Principles} introduced a lumped zero-dimensional LLS model that has the potential of describing DPZ trajectories (i.e., temporal evolution of DPZ under dynamic operating conditions), yet experimental validation is missing due to lack of suitable data. Both steady-state and lumped models have shown less than 25\% accuracy in predicting flow rates at which flooding occurs \citep{Zhai.2025}. These shortcomings highlight the need for alternative approaches, such as data-driven modeling. However, purely data-driven methods typically require extensive amounts of data to generalize reliably \citep{shlezinger2022modelbaseddeeplearning}. 

Physics-informed neural networks (PINNs) \citep{Raissi2019Physics-informedEquations} combine first-principle and data-driven modeling by embedding known physical relationships within the data-driven training process. Thus, PINNs can generalize from scarce training data while adhering to the governing physics. Our previous work \citep{VELIOGLU2025108899} presented an \emph{in-silico} proof of concept in which mechanistic model data were used as ground truth to demonstrate that a PINN trained with limited data and incomplete physics can predict DPZ height over short time intervals. However, estimation of DPZ height from experimental settler data for longer time intervals remains unexplored.

One possible strategy to further reduce the demand for experimental data is to first train PINN models on physics equations and/or synthetic data generated from a low-fidelity (i.e., approximate) mechanistic model, and then fine-tune them using high-fidelity data. \citet{Chakraborty2021} showed that a two-stage training procedure for PINNs involving a low-fidelity physics model for pretraining and high-fidelity data for fine-tuning can improve prediction accuracy when the known governing physics equations are only approximate and the high-fidelity data is scarce. \citet{Wang2025TL} compared different transfer learning schemes for PINNs, including full fine-tuning and parameter-efficient updates, to adapt PINNs across different tasks. \citet{Mustajab2024} have examined high-frequency and multi-scale problems, where pretraining on baseline low-frequency models provides stable initialization that enhances PINN accuracy when predicting high-frequency problems. \citet{Prantikos2023} proposed selecting pretraining datasets based on domain similarity for nuclear reactor applications, illustrating how task relevance in PINNs accelerates the training in the fine-tuning step. Related developments outside the PINN literature also point to the value of combining simulation-based pretraining with experimental data, for example in process control of froth flotation systems under partial observability with Gaussian processes \citep{Wang2025Gaussian}, or in reinforcement learning where sample efficiency is critical \citep{sim2real-RL}.

While the PINN provides a predictive model of the system dynamics, it cannot by itself reconstruct the true system state under realistic operating conditions, where phase heights are not measurable, initial conditions are unknown, and flow-rate measurements are noisy. Therefore, an estimation scheme is required to fuse model predictions with measurements and enable real-time phase height estimation. To this end, \citet{Arnold2021StatespaceNetworks} showed that PINNs can be effectively integrated within an Extended Kalman Filter (EKF) \citep{Gelb:1974}, since Jacobians can be computed through automatic differentiation \citep{naumannAD}. \citet{Cuomo2024} combined a PINN with the EKF to predict structural displacements in railway motion. In vehicle applications, \citet{Tan2023} combined a PINN with the Unscented Kalman Filter (UKF) \citep{JulierUKF} to estimate vehicle attitude, velocity and position with increased accuracy, while \citet{decurto2024hybrid} demonstrated the use of a PINN-based predictive model as state-transition model in UKF for dynamic models such as a double pendulum system. \citet{Fang2025} fine-tuned a pretrained dynamic model with physics guidance to represent hybrid dynamics before embedding them in an EKF framework for vehicle state estimation. These examples suggest that PINNs can serve as surrogates for predictive models when governing equations are incomplete, while the filtering step accounts for measurement noise and model uncertainty. To complement this, ensemble learning \citep{Breiman1996StackedRegressions, dietterich2000ensemble} constitutes an additional technique for improving prediction robustness and uncertainty quantification.

Although PINNs have been combined with Kalman-type filters for state estimation, their application for estimating DPZ heights in pilot-scale LLS has not yet been investigated. Demonstrating that DPZ height can be estimated from readily available volume flow-rate measurements with the aid of a PINN would provide significant benefits for LLS operation.

This work aims to estimate the phase heights in a pilot-scale LLS. The considered system is a horizontal cylindrical gravity separator (length = 1 m, radius = 0.1 m), where the dispersion flows along the axial direction and separates into a light phase, a dispersion layer (DPZ), and a heavy phase. We conducted step-response experiments in the pilot-scale LLS by varying the volume flow rate. These experiments capture the transient behavior of DPZ heights and provide data for training, validation, and testing (\cref{sec:settler}). The low-fidelity mechanistic LLS model (\cref{sec:mm_lls}) from our previous work \citep{VELIOGLU2025108899} combines simple volume balance equations with complex droplet coalescence and sedimentation submodels. The model is only approximate, as it assumes constant phase heights along the separator length (band-shaped DPZ), neglects axial (convective) transport of the DPZ, droplet–droplet coalescence, and axial and vertical variations in Sauter mean diameter. We generate synthetic pretraining data using the mechanistic model, but use only the volume balance equations as physics regularization in the PINN, as the droplet coalescence and sedimentation submodels are computationally challenging to implement in the PINN due to the necessary discretization of the axial length and droplet classes. A PINN model of the LLS dynamics is then developed by pretraining on synthetic data from the low-fidelity mechanistic model and then fine-tuning with experimental phase-height and flow-rate data (\cref{sec:method}), similar to the training strategy from \citet{Chakraborty2021}. The trained PINN is subsequently integrated into an EKF-inspired state estimator to estimate phase heights solely from volume flow-rate measurements during deployment (\cref{sec:results}). Conclusions and an outlook on future work are presented in \cref{sec:conclusion}.

This work integrates elements from previous studies across different domains, including our earlier works on the experimental LLS setup \citep{Zhai.2025}, the mechanistic LLS model \citep{VELIOGLU2025108899}, and the PINN model that is largely based on \citet{VELIOGLU2025108899}. Moreover, we describe the established EKF method \citep{Gelb:1974}. For self-containment and readability, relevant portions of these works are repeated here with appropriate citation.

\section{Experimental setup} \label{sec:settler}

\subsection{Materials and experimental liquid-liquid separator setup}
\begin{figure}[htb]
    \centering
    \includegraphics[width=1.0\linewidth]{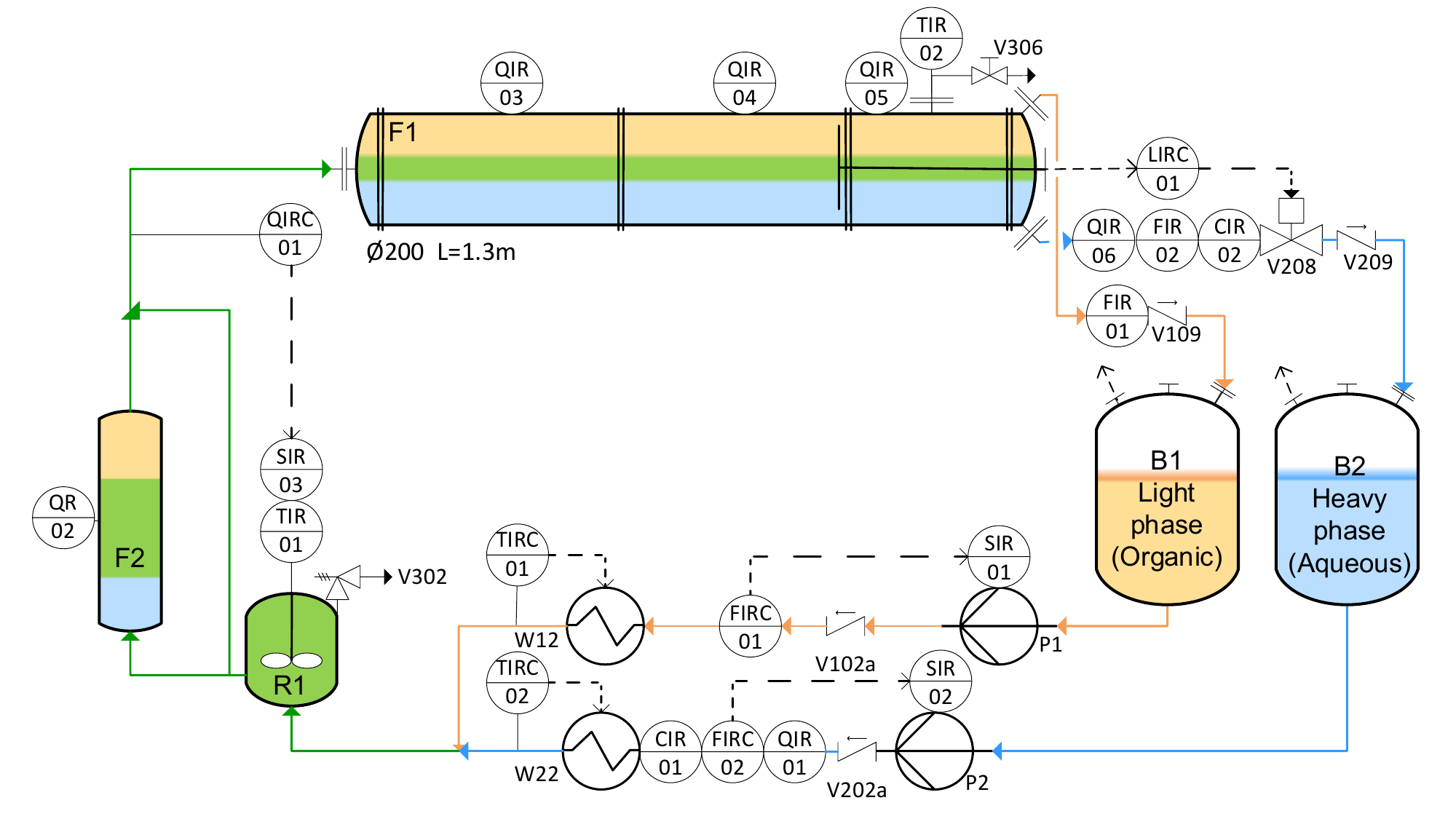}
    \caption{Piping and instrumentation diagram of the experimental setup with DN 200 separator in the RWTH Aachen Fluid Process Engineering (AVT.FVT) lab. Blue, orange, and green colors mark the aqueous, organic, and dispersion phases (DPZ). Reproduced from \citet{Zhai.2025}.}
    \label{fig:PID-diagram}
\end{figure}

All experiments are conducted in the pilot-scale continuous LLS at \SI{30}{\celsius} shown in \cref{fig:PID-diagram}. 1-Octanol (99 \%, Häberle Labortechnik GmbH \& Co. KG) is dispersed in deionized water (conductivity \SI{13}{\micro\siemens\per\centi\meter}). Physical properties of the saturated liquids at \SI{30}{\celsius} are listed in \cref{tab:physicalproperties}. 
The effective length of the separator is set to \SI{1}{\meter} by a weir. The dispersion is generated in the stirred tank R1, inlet volume flows are controlled by frequency-controlled peripheral pumps P1 and P2, temperature is controlled by the heater W12 and W22, and the interfacial position of the separator is controlled by the check valve V208 which manipulates the heavy phase outlet volume flow. Outlet and inlet streams of the separator are recycled by the storage vessels B1 and B2. Ingoing and outgoing flow rates of the aqueous and organic phase are recorded by installed Coriolis flow meters FIRC01, FIRC02, FIR01, FIR02. Heights of the DPZ were measured by external cameras along the separator QIR03, QIR04, QIR05.
A complete list of used apparatuses and measurement devices can be found in \citet{Zhai.2025}.

\begin{table}[htb]
    \centering
    \caption{Physical properties of the saturated biphasic plant system at \SI{30}{\celsius}.}
    \begin{tabular}{@{}lccc@{}}
    \toprule
              & Density $\rho $ & Dynamic viscosity $\eta  $ & Interfacial tension $\gamma $\\ \midrule
    water     & 996 \si{\kilo\gram\per\cubic\meter}     & 0.82 \si{\milli\pascal\second}             & \multirow{2}{*}{8.2 \si{\milli\newton\per\meter}}   \\
    1-octanol & 825 \si{\kilo\gram\per\cubic\meter}     & 6.04 \si{\milli\pascal\second}            &   \\ \bottomrule
    \end{tabular}
    \label{tab:physicalproperties}
\end{table}

\subsection{Optical measurement of the dense-packed zone height}\label{sec:methods-opticalmeasurement}
\begin{figure}
    \centering
    \includegraphics[width=1.0\linewidth]{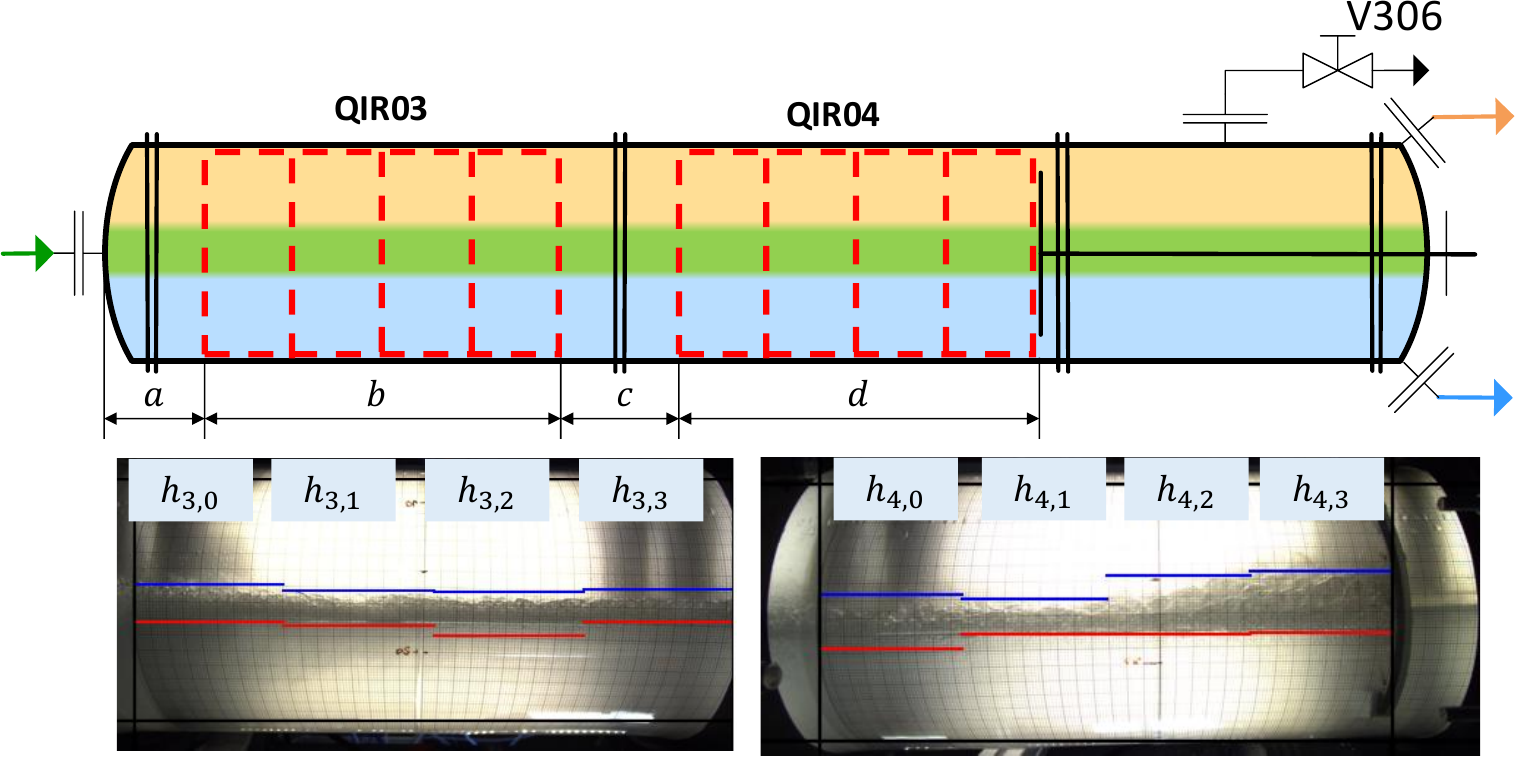}
    \caption{Detection of DPZ heights along the separator with effective length of \SI{1}{\meter}. Images refer to detections in QIR03 (left) and QIR04 (right). In each image, four heights are detected for same widths. The distances between inlet and first height detection, width of QIR03, width between QIR03 and QIR04, and width of QIR04 is $a=\SI{21}{\centi\meter}$, $b=\SI{36}{\centi\meter}$, $c=\SI{10}{\centi\meter}$, $d=\SI{33}{\centi\meter}$, respectively.}
    \label{fig:Separator_CamPositions}
\end{figure}

\begin{figure}[htbp]
    \centering
    
    \includegraphics[width=1.0\linewidth]{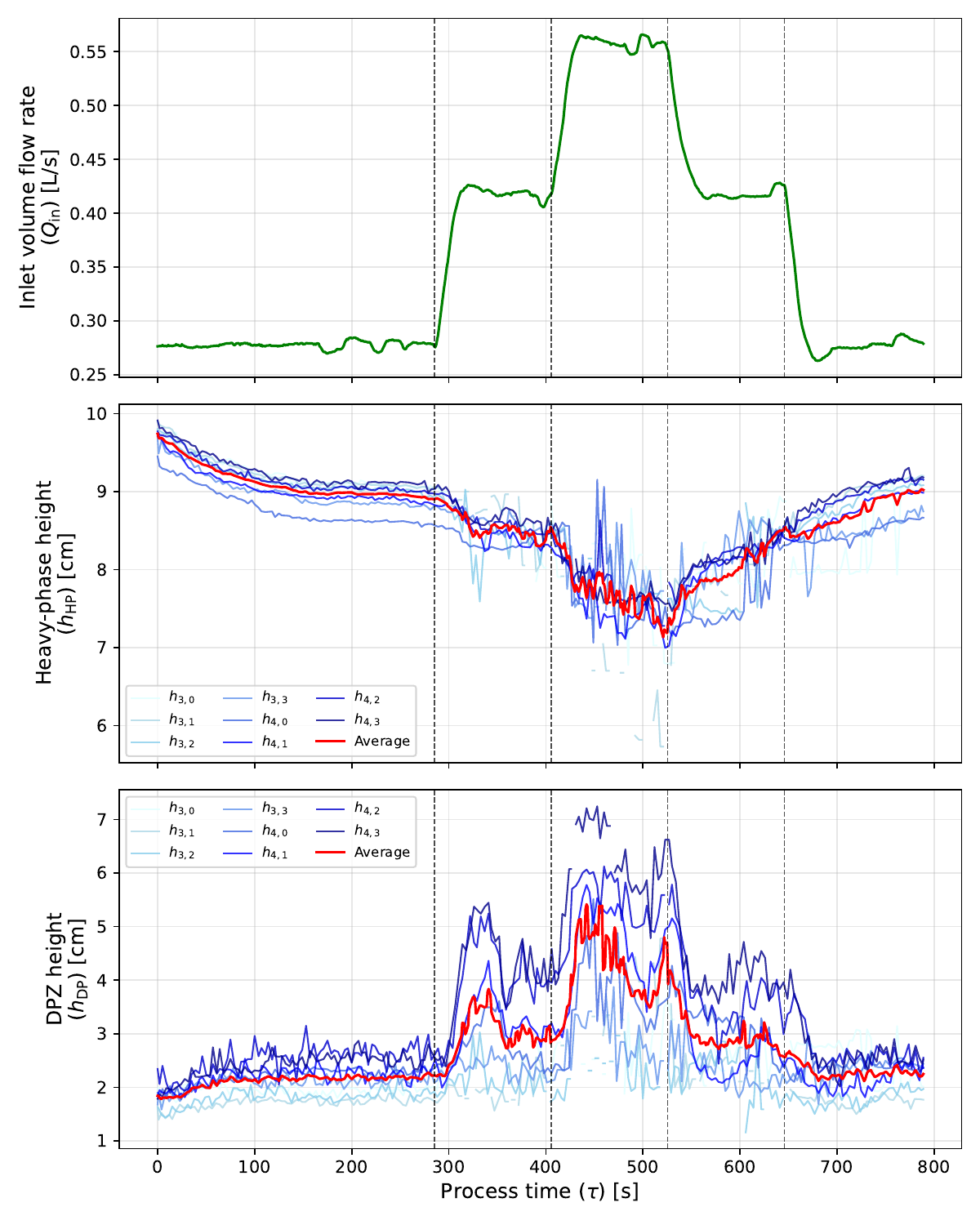}
    
    \caption{Pre-processed inlet volume flow rate and phase height measurements for the \textbf{training trajectory}, measured at eight different positions at \textbf{QIR03} and \textbf{QIR04} (see \cref{fig:Separator_CamPositions}). Vertical dashed lines denote the times of inlet volume flow setpoint changes.}
    \label{fig:trajectory1_preprocessed}
\end{figure}

The phase heights for both the heavy (water) phase and the DPZ are determined from two camera recordings (cf. QIR03, QIR04 in \cref{fig:PID-diagram}) taken at four distinct axial positions from each camera. The exact detection positions are shown in \cref{fig:Separator_CamPositions}. \cref{fig:trajectory1_preprocessed} shows the optically measured phase heights for the training trajectory, where the quantities of interest are the average heavy-phase and DPZ heights over the available detection positions; these averaged phase heights are the variables estimated in the present study.  

Similar to the detection algorithm for detecting drop size distributions in endoscope images \citep{Sibirtsev.2023,Sibirtsev.2024} or external camera images \citep{Palmtag.2025}, heights are detected by the convolutional neural network based object detection algorithm YOLOv8 trained on manually labelled lab data \citep{Redmon.2016}. The manual labelling itself contains errors due to the uneven shape of the drops in the DPZ. Moreover, detection errors can occur due to illumination conditions, often appearing as spikes in the recorded trajectories (cf. \cref{fig:trajectory1_preprocessed}). Furthermore, occasional missing measurements, resulting from object detection errors, appear as discontinuities in the recorded trajectories (cf. \cref{fig:trajectory1_preprocessed}). An exemplary detection is shown in \cref{fig:Separator_CamPositions}.

\subsection{Design of experiments for separator dynamics} \label{sec:exp-design}

The experiments are conducted to investigate the dynamic accumulation of the DPZ phase in the pilot-scale LLS under variations in total volumetric flow rate. Previous studies \citep{Zhai.2025, Ye.2023b, PADILLA1996267, Jeelani1988} identified the flow rate as the dominant operating parameter. The flow rate can equivalently be expressed in terms of the theoretical residence time in the separator, defined as $\tau_{\mathrm{res}} = V_{\mathrm{sep}} / Q_{\mathrm{in}}$, which represents the time available for droplet coalescence and sedimentation and thus governs DPZ formation. All other conditions, including temperature, stirrer speed, feed phase fraction, and interfacial position, are kept constant at \SI{30}{\celsius}, \SI{600}{rpm}, \SI{0.5}{}, and \SI{100}{\milli\meter}, respectively. System input is provided as stepwise changes in flow rate, with both the temporal step size (duration of constant operation) and the flow-rate step size (magnitude of change) varied across trajectories. One trajectory extends the flow-rate range beyond that of the others and is used to assess the extrapolation capability of the model in that regime. A practical example in an industrial LLS could be upstream disturbances causing unusually high or low flow rates. Note that the findings cannot be generalized to more extreme variations in flow rates or other types of extrapolation, e.g., changes in phase fraction or changes in temperature. 

In total, four distinct trajectories were acquired, with detailed step durations and flow-rate levels summarized in \cref{tab:designofexperiments}. Before each experiment, the plant was heated and circulated for at least one hour until homogeneous conditions were established. Each trajectory was measured only once as repeated experiments under identical conditions would have required repeating the time-intensive preconditioning of the pilot-scale setup. This results in a limited and noisy dataset, which is addressed by the proposed framework in \cref{sec:method,sec:results}. Specifically, data scarcity is mitigated through physics-based regularization and pre-training on synthetic data (cf. \cref{sec:training,sec:physics_residuals}), while ensemble learning (cf. \cref{sec:validation}) accounts for model uncertainty, and filtering (cf. \cref{sec:EKF}) accounts for both model uncertainty and measurement noise. The use of single-run experimental data therefore not only reflects practical constraints, but it is intended to demonstrate that the proposed modeling approach can achieve generalization under data-lean scenarios.

\begin{table}[htbp]
    \centering
    \caption{Summary of setpoints for the experimental trajectories.}
    \resizebox{\textwidth}{!}{%
    \begin{tabular}{l l llllllllll}
    \toprule
    \textbf{Trajectory} & \textbf{Time between} & \multicolumn{10}{c}{\textbf{Volume flow rate $Q_{\textrm{in}}$/ \si{\liter\per\second}}} \\
    \cmidrule(lr){3-12}
     & {\bf step change} & 1 & 2 & 3 & 4 & 5 & 6 & 7 & 8 & 9 & 10 \\
    \midrule
    1 -- training          & 2.00 \si{\minute}   & 0.278 & 0.417 & 0.556 & 0.417 & 0.278 &      &      &      &      &      \\
    2 -- validation        & 1.50 \si{\minute}   & 0.278 & 0.417 & 0.556 & 0.417 & 0.278 &      &      &      &      &      \\
    3 -- test (interpolation)  & 2.00 \si{\minute}  & 0.278 & 0.347 & 0.417 & 0.486 & 0.556 & 0.486 & 0.417 & 0.347 & 0.278 &      \\
    4 -- test (extrapolation)  & 2.00 \si{\minute}  & 0.208 & 0.347 & 0.486 & 0.625 & 0.486 & 0.347 & 0.208 &      &      &      \\
    \bottomrule
    \end{tabular}
    }
    \label{tab:designofexperiments}
\end{table}

\subsection{Post-processing of phase height measurements}
\label{sec:exp-postproc}

Our PINN modeling approach (cf. \cref{sec:mm_lls}) assumes a band-shaped operation, meaning that phase heights would be constant along the separator length. This assumption reduces the separator to a lumped system by eliminating axial height variations, effectively replacing a PDE-based description with an ODE system. Such lumped models are conceptually and computationally easier to handle. Although the band-shaped assumption is violated at higher volume flow rates, as evidenced by the wedge-shape DPZ formed at QIR04 in \cref{fig:Separator_CamPositions}, we use the mean of the available measurements to represent the average DPZ height to be modeled and train a separate neural network to estimate the DPZ height at the separator outlet, which is most critical for flooding, based on the average DPZ height (cf. \cref{sec:dpz_end}).

Object detection based phase height readings arrive at irregular intervals, typically every 2–3 seconds; we therefore linearly interpolate the phase height measurements to 1-second resolution to match the resolution of the flow-rate measurements. 

\begin{figure}[htbp]
    \centering
    
    \begin{subfigure}[b]{0.41\textwidth}
        \centering
        \includegraphics[width=\linewidth]{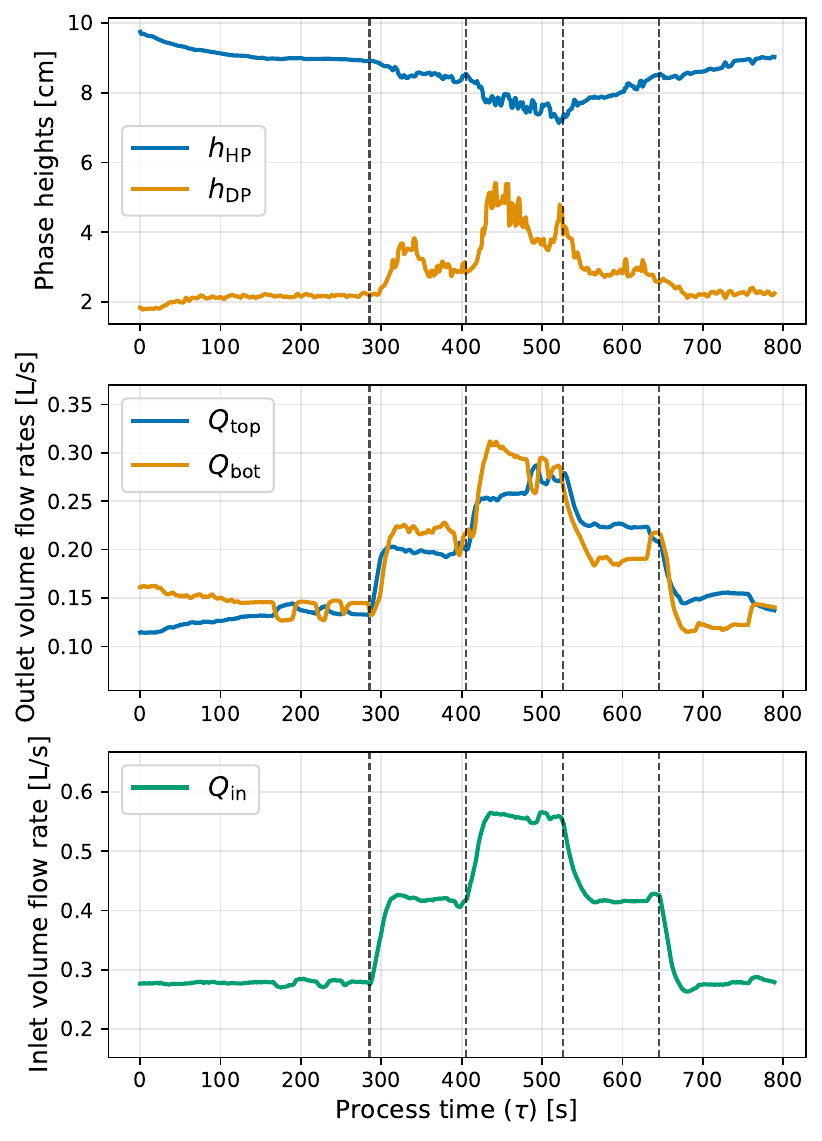}
        \caption{Training}
        \label{fig:traj_1}
    \end{subfigure}
    \begin{subfigure}[b]{0.41\textwidth}
        \centering
        \includegraphics[width=\linewidth]{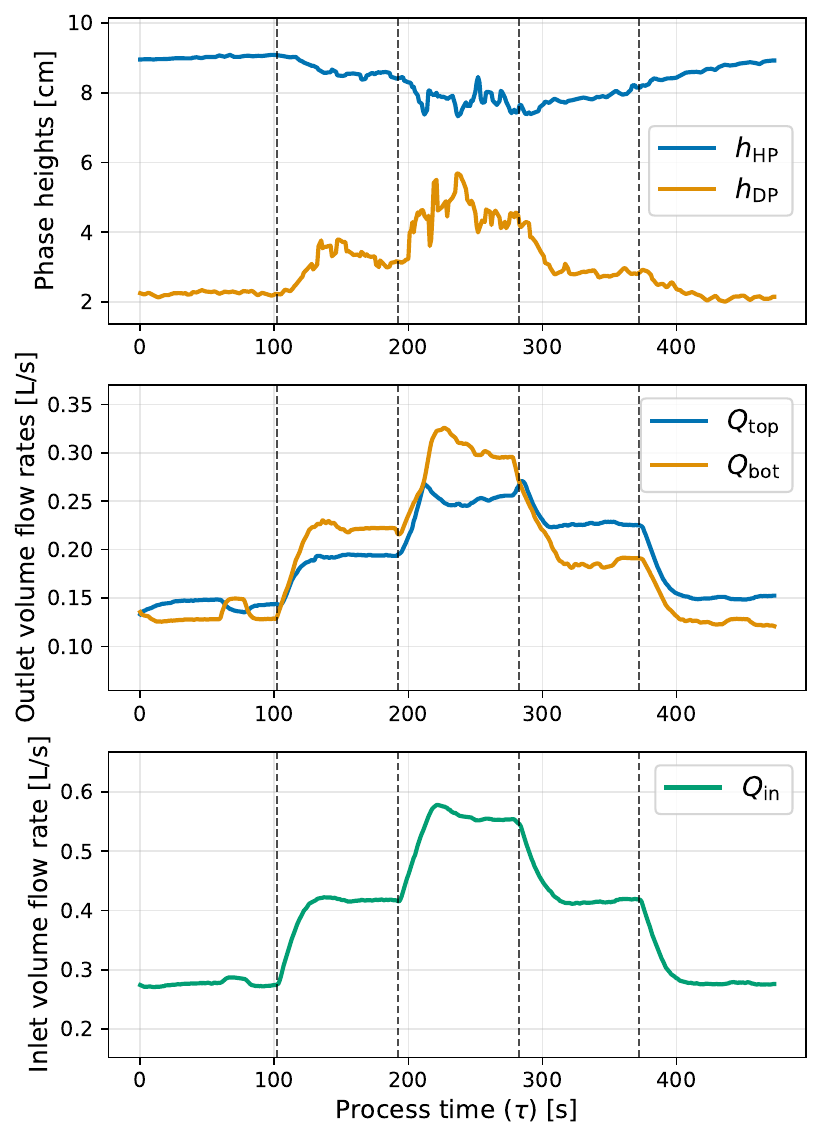}
        \caption{Validation}
        \label{fig:traj_2}
    \end{subfigure}

    \vspace{1em}

    \begin{subfigure}[b]{0.41\textwidth}
        \centering
        \includegraphics[width=\linewidth]{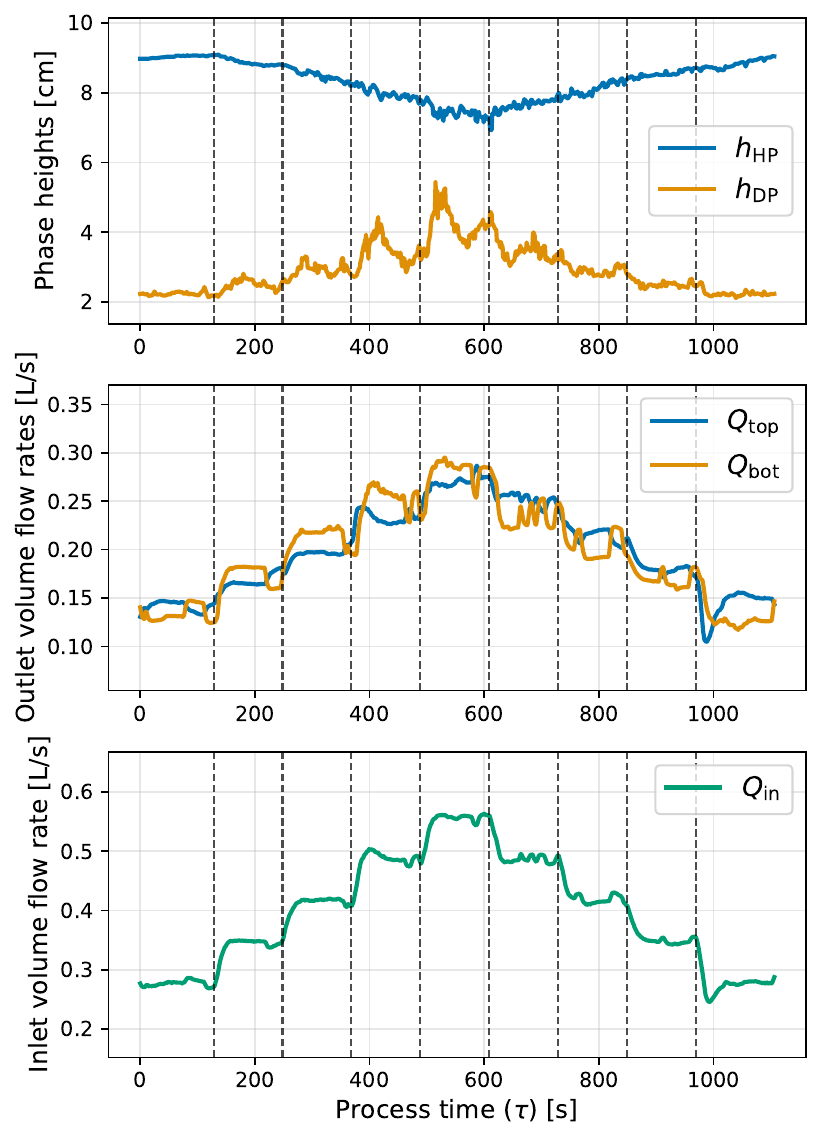}
        \caption{Interpolation test}
        \label{fig:traj_3}
    \end{subfigure}
    \begin{subfigure}[b]{0.41\textwidth}
        \centering
        \includegraphics[width=\linewidth]{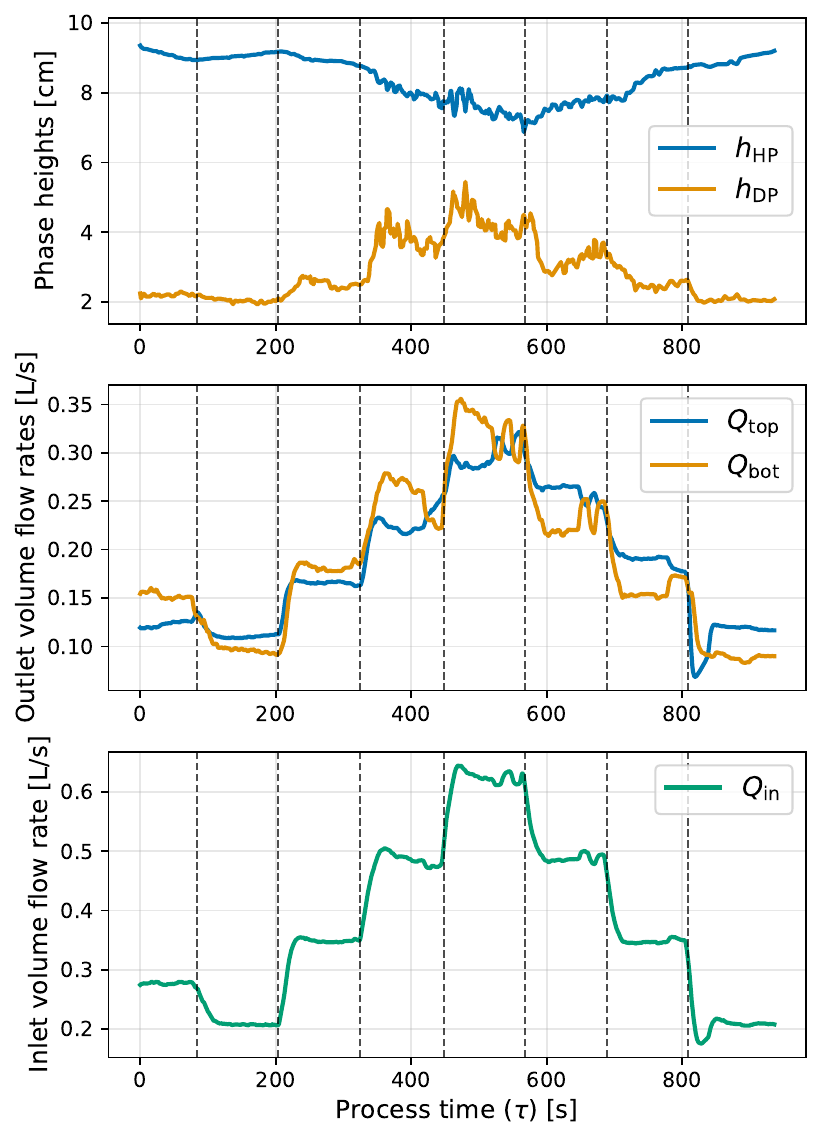}
        \caption{Extrapolation test}
        \label{fig:traj_4}
    \end{subfigure}
    
    \caption{Post-processed experimental trajectories obtained from the gravity settler. The quantities $\hw$ and $\hp$ denote the average heavy-phase height and DPZ height, respectively, computed over the eight detection positions. $\Qin$ denotes the inlet volume flow rate, and $\Qb$ and $\Qt$ denote the bottom and top outlet volume flow rates, respectively. Vertical dashed lines denote the times of inlet volume flow setpoint changes.}
    \label{fig:experimental_trajectories}
\end{figure}

We acquired four trajectories with varying time duration from the experimental gravity settler setup, as demonstrated in \cref{tab:designofexperiments}. From those, we use trajectory \#1 (\cref{fig:traj_1}) for training, trajectory \#2 (\cref{fig:traj_2}) for validation, and trajectories \#3 and \#4 for testing. Specifically, trajectory \#3 (\cref{fig:traj_3}) is used for the interpolation test, as it contains inlet volume flow rates within the range $Q_{\textrm{in}} \in [0.278,\,0.556]\,\si{\liter\per\second}$, but has values at intermediate levels not encountered during training (e.g., $0.347\,\si{\liter\per\second}$ in \cref{tab:designofexperiments}). Trajectory \#4 (\cref{fig:traj_4}) is used for the extrapolation test, as it includes inlet volume flow rates $Q_{\textrm{in}}$ outside the range observed in the training trajectory (e.g., $0.625\,\si{\liter\per\second}$ in \cref{tab:designofexperiments}). The validation trajectory is used in hyperparameter tuning and tuning for the EKF application (cf. \cref{sec:EKF}). The respective bounds for all variables are given in \cref{tab:bounds}. The resulting experimental dataset is denoted by $\mathcal{D}_{\mathrm{exp}}$.

\begin{table}
\centering
\caption{Operating ranges for the phase heights and volume flow rates. The lower bound is denoted by $\textrm{lb}$, and the upper bound is denoted by $\textrm{ub}$.}
\begin{tabular}{@{}cccccccc@{}}\toprule
\multirow{2}{*}{$\textrm{Variable}$} & \multirow{2}{*}{$\textrm{Unit}$} & \phantom{a}& \multicolumn{2}{c}{\textrm{Interpolation bounds}} & \phantom{a}& \multicolumn{2}{c}{\textrm{Extrapolation bounds}} \\
\cmidrule{4-5} \cmidrule{7-8}
& && $\textrm{lb}$ & $\textrm{ub}$ && $\textrm{lb}$ & $\textrm{ub}$ \\ \midrule
$\hw$ & \unit{\meter} && 0.071  & 0.091 && 0.067 & 0.100\\
$\hp$ & \unit{\meter} && 0.023 & 0.059 && 0.019 & 0.069\\
$\Qin$ & \unit{\liter\per\second} && 0.245 & 0.563 && 0.175  & 0.644 \\
$\Qt$ & \unit{\liter\per\second} && 0.105 & 0.286 && 0.069 & 0.321 \\
$\Qb$ & \unit{\liter\per\second} && 0.117 &  0.295 && 0.083 &  0.356 \\
\bottomrule
\end{tabular}
\label{tab:bounds}
\end{table}

\section{Low-fidelity mechanistic LLS model} \label{sec:mm_lls}
\begin{figure}[htb]
    \centering
    \includegraphics[width=0.66\linewidth]{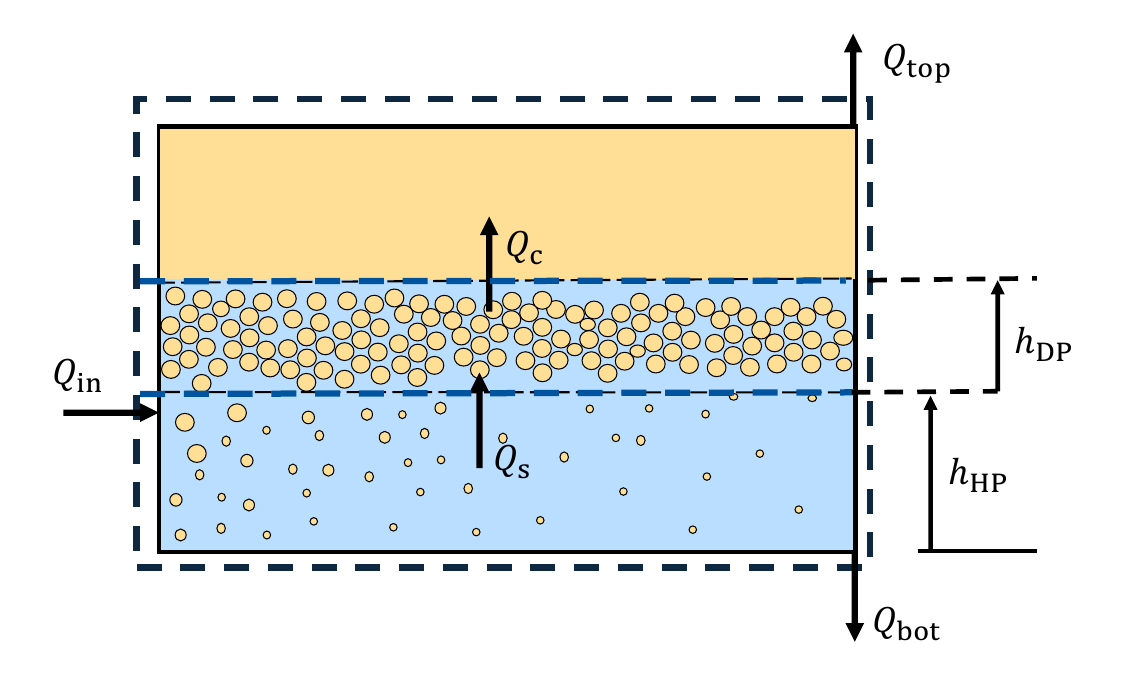}
    \caption{Mechanistic model schematic of the pilot-scale liquid–liquid separator. Adapted from \citet{VELIOGLU2025108899}.}
    \label{fig:LLS-MM}
\end{figure}

The low-fidelity mechanistic LLS model was introduced in \citet{VELIOGLU2025108899} and is visualized in \cref{fig:LLS-MM}. It divides the settler content into three zones: the light phase, the DPZ, and the heavy phase. The geometry of the entire fluid is defined by an effective length of $L = \SI{1}{\meter}$ and a radius of $r = \SI{0.1}{\meter}$. The total inlet flow rate, $\Qin$, enters from the left with a Sauter mean diameter $d_\mathrm{32,in}$ and organic phase fraction of $\epsilon_{\textrm{in}} = 0.5$, and is assumed to disperse completely into the heavy phase. The inlet drop size distribution is modeled by the self-similarity assumption with a normalized standard deviation of the volume-based log-normal drop size distribution $\sigma_{\mathrm{selfsimilar}} = \sigma/d_{\mathrm{32,in}} = 0.32$ \citep{Kraume.2004,Ye.2023}. 

The DPZ is assumed to be band-shaped, meaning constant phase heights, $\hw$ (heavy phase) and $\hp$ (DPZ), along the separator length. Within the DPZ, droplet-droplet coalescence is assumed to be negligible, while droplet-interface coalescence results in a volume flow $Q_{\mathrm{c}}(t)$ into the light phase. The dispersed droplets in the heavy phase rise as a droplet swarm with swarm exponent $n_{\mathrm{swarm}}=2$ \citep{Mersmann.1980,Kampwerth.2020,Ye.2023b}, resulting in a volume flow $Q_{\mathrm{s}}(t)$ into the DPZ. Both the coalescence $Q_{\mathrm{c}}(t)$ and sedimentation $Q_{\mathrm{s}}(t)$ flow rates are derived from complex submodels that discretize the axial length of the separator into 200 spatial discretization elements and the droplet size distribution into 50 droplet size classes. Additionally, the volume flow of water from the heavy phase into the DPZ is expressed by the coalescence rate $Q_{\mathrm{c}}(t)$ and the sedimentation rate $Q_{\mathrm{s}}(t)$, by assuming a constant DPZ hold-up of $\bar{\epsilon}_\mathrm{DP} = 0.9$ in accordance with the separator model proposed by \citet{Henschke.1995}. The convective transport of the DPZ is also neglected, as well as axial variation of the Sauter mean diameter and its effect on the DPZ.

The separated light and heavy phases exit with outlet volume flow rates $\Qt$ and $\Qb$, respectively, both assumed to be free of entrained droplets. The total volume of the three phases is kept constant and equal to the separator volume, yielding the volume balance given by \cref{eq:settler-1}. The mechanistic LLS model takes as inputs the initial phase heights $\hw(t=0)$ and $\hp(t=0)$, the inlet volume flow rate $\Qin$, the bottom volume flow rate $\Qb$, the Sauter mean diameter at inlet $d_\mathrm{{32,in}}$, and provides as outputs the phase heights $\hw(t)$, $\hp(t)$, the top outlet volume flow rate $\Qt(t)$, the coalescence rate $Q_\mathrm{c}(t)$ and sedimentation rate $Q_\mathrm{s}(t)$. Further details of the mechanistic LLS model can be found in \cite{VELIOGLU2025108899}.

The mechanistic LLS model \cite{VELIOGLU2025108899} is based on the following volume balance equations obtained after transforming the volume of a cylindrical segment to the height of each segment, similar to \citet{Backi2018AFirst-Principles}:
\begin{subequations}
    \label{eq:settler}
    \begin{align}
        \label{eq:settler-1} 0 &= g_{\mathrm{sep}}(\Qin(t), \Qb(t), \Qt(t)) = \Qin(t) - \Qb(t) - \Qt(t), \\
        \begin{split}
                \label{eq:settler-2} \hpd(t) &= f_{\mathrm{DP}}(\hw(t), \hp(t), \Qin(t), \Qb(t), Q_{\mathrm{c}}(t), Q_{\mathrm{s}}(t)) \\ &= \frac{\Qin(t) - \Qb(t) - Q_{\mathrm{c}}(t) }{2L\sqrt{\left(\hp(t) + \hw(t)\right)\left(2r-\hp(t)-\hw(t)\right)}} \\ & - \frac{\Qin(t)  -  \Qb(t)  -  Q_{\mathrm{s}}(t) \frac{1}{\bar{\epsilon}_\mathrm{DP}} +  Q_{\mathrm{c}}(t) \frac{1-\bar{\epsilon}_\mathrm{DP}}{\bar{\epsilon}_\mathrm{DP}} }{2L\sqrt{\hw(t)(2r-\hw(t))}},
        \end{split} \\
        \begin{split}
                \label{eq:settler-3}\hwd(t) &= f_{\mathrm{HP}}(\hw(t), \hp(t), \Qin(t), \Qb(t), Q_{\mathrm{c}}(t), Q_{\mathrm{s}}(t)) \\ &=  \frac{\Qin(t)  -  \Qb(t)  -  Q_{\mathrm{s}}(t) \frac{1}{\bar{\epsilon}_\mathrm{DP}} +  Q_{\mathrm{c}}(t) \frac{1-\bar{\epsilon}_\mathrm{DP}}{\bar{\epsilon}_\mathrm{DP}} }{2L\sqrt{\hw(t)(2r-\hw(t))}},            
        \end{split}
    \end{align}
\end{subequations}
Note that the submodels governing the coalescence and sedimentation rates, $Q_{\mathrm{c}}(t)$ and $Q_{\mathrm{s}}(t)$, are not included here due to their complexity; the complete formulations are provided in the Supplementary Materials (SM4) of \citet{VELIOGLU2025108899}. Their dependencies on the states and parameters are expressed as:
\begin{subequations}
\label{eq:submodels}
\begin{align}    
    \begin{split}
        \label{eq:settler-4} Q_{\mathrm{s}}(t) &= g_{\mathrm{s}}(\hw(t), d_{\mathrm{32,in}}, \sigma_{\mathrm{selfsimilar}}, n_{\mathrm{swarm}}, \epsilon_{\mathrm{in}}, \Delta\rho, \eta_{\mathrm{HP}}, r, L), 
    \end{split} \\
    \begin{split}
        \label{eq:settler-5} Q_{\mathrm{c}}(t) &= g_{\mathrm{c}}(\hw(t), \hp(t), d_{\mathrm{32,in}}, \sigma_{\mathrm{selfsimilar}}, n_{\mathrm{swarm}}, \epsilon_{\mathrm{in}} ,\Delta\rho, \eta_{\mathrm{HP}}, \gamma, r, L)
    \end{split}
\end{align}    
\end{subequations}
where $\Delta\rho$ is the density difference between the two liquids,  $\eta_{\mathrm{HP}}$ is the dynamic viscosity of the heavy phase and $\gamma$ is the interfacial tension (cf. \cref{tab:physicalproperties}).

Using the mechanistic LLS model described above, we generate 1000 segments with a duration of $\SI{1}{\s}$ where we sample the model inputs with Latin Hypercube Sampling (LHS) \citep{Iman1981AnAssessment} from the bounds provided in \cref{tab:bounds}. We use all of the segments as pretraining data. The simulation segments have a constant step size of $\Delta t= \SI{0.1}{\second}$ for the PINN (cf. \cref{sec:method}). The pretraining dataset is referred to as $\mathcal{D}_{\mathrm{sim}}$. We choose the bounds for initial states and control inputs that correspond to minima and maxima of state values in the experimental trajectories (cf. \cref{tab:bounds}), including the extrapolation trajectory. The predictive capability of the low-fidelity mechanistic model is later evaluated by forward simulation against experimental trajectories, alongside the proposed PINN approach (see \cref{sec:validation}).

\section{Physics-informed neural network model of the settler dynamics}\label{sec:method}

In this section, we present the dynamic modeling of the gravity settler through the PINN approach. In \cref{sec:arch}, we present the PINN architecture. The physics equations and the associated modeling assumptions are presented in \cref{sec:physics_residuals} and data handling for the PINN training is presented in \cref{sec:data_handling}. \cref{sec:training} discusses the PINN training strategy, \cref{sec:VNN} presents the purely data-driven neural network for comparison, and \cref{sec:validation} presents prediction results and comparison of different models. 

The PINN model used in this work is largely based on our LLS PINN model introduced in \citet{VELIOGLU2025108899}. We employ the same network architecture (a feedforward neural network), activation functions, two-optimizer training scheme and the dynamic weighting scheme. For the sake of self-containment, we restate these components in this section. The notable changes introduced in this work are as follows:
(i) In contrast to the previous study, which used exclusively synthetic simulation data, we now also use experimental data for both training and testing.
(ii) The outlet volume flow rates, $\Qb$ and $\Qt$, are treated as PINN outputs instead of PINN inputs, whereas the inlet volume flow rate, $\Qin$, is treated as a PINN input (control action) instead of a PINN output.
(iii) A shorter PINN time horizon is used to match the resolution of the experimental data.
(iv) The Sauter mean diameter at the inlet, $d_{\mathrm{32,in}}$, and the coalescence parameter, $r_v$, are no longer used as inputs; the former did not contribute substantial predictive capability, and the latter depends primarily on temperature, which is held constant throughout this work.
(v) Phase heights are no longer modeled from the bottom of the separator.
(vi) Loss terms are conceptually the same, although exact terms are different due to changes mentioned in points (ii) and (v).
(vii) A new two-stage training scheme is introduced (pretraining and fine-tuning).

\subsection{Model architecture} \label{sec:arch}

We define a PINN with learnable parameters $\bm{\theta}$ that takes PINN time $t \in [0, T]$, initial heights $\Vx(0) = [\hp(0), \hw(0)]^\top$ and inlet volume flow (control) $\Vu = [\Qin]$ as input and outputs the measurable settler heights $\Vx(t) = [\hp(t), \hw(t)]^\top$, measurable outlet volume flows $\Vy(t) = [\Qb(t), \Qt(t)]^\top$ and immeasurable internal flows $\Vz(t) = [Q_{\mathrm{c}}(t), Q_{\mathrm{s}}(t)]^\top$:

\begin{equation}
    [\hat{h}_{\mathrm{HP}}(t), \hat{h}_{\mathrm{DP}}(t), \hat{Q}_{\mathrm{bot}}(t), \hat{Q}_{\mathrm{top}}(t), \hat{Q}_c(t), \hat{Q}_s(t)]^\top = \mathrm{PINN}_{\bm{\theta}}(t, \hw(0), \hp(0), \Qin)
\end{equation}

\begin{figure}
    \centering
    \includegraphics[scale=0.5]{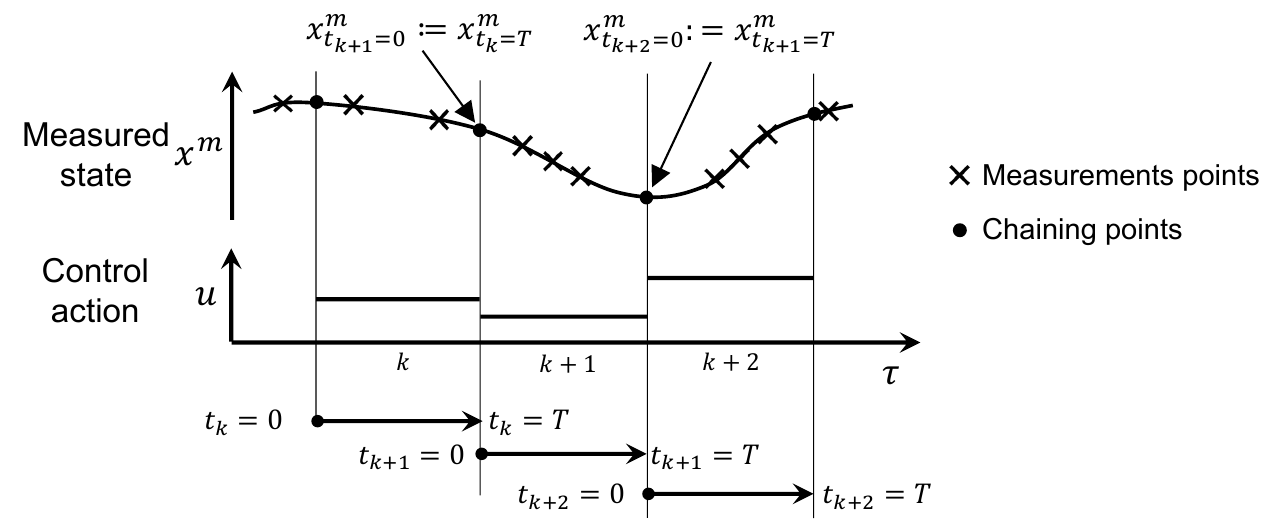}
    \caption{Relationship between PINN time $t$ and process time $\tau$. Reproduced from \citet{VELIOGLU2025108899}.}
    \label{fig:step-wise-control}
\end{figure}

The hat ($\hat{~}$) notation denotes a PINN prediction. Note that the PINN time $t \in [0, T]$ differs from the process time $\tau$. Specifically, we partition the process time domain of a \emph{trajectory} into shorter \emph{segments} of duration $T = \SI{1}{\second}$, which corresponds to the length of PINN time $t \in [0, T]$ (cf. Fig \ref{fig:step-wise-control}) \citep{VELIOGLU2025108899}. This segmentation keeps the control actions constant within each segment.

The given input configuration allows the PINN model to be trained to extensively cover the state and control action spaces. The PINN can be used for multi-step predictions by iteratively feeding the predicted final state of one segment as the initial condition for the next segment (forward simulation, cf. \cref{fig:step-wise-control}).

\begin{figure}
    \centering
    \includegraphics[scale=0.42]{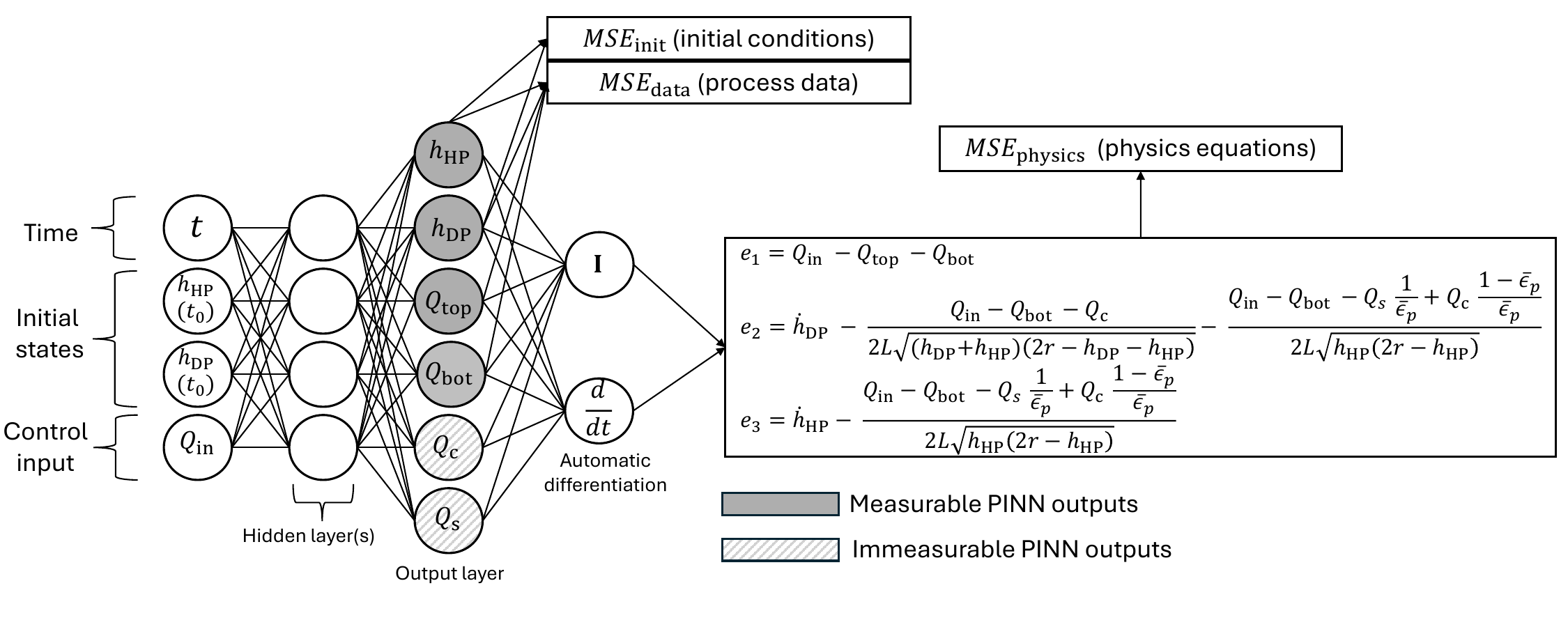}
    \caption{Network schematic of the PINN model for liquid-liquid separator. Note that the figure does not show the actual depth and width of the hidden layers. Time dependence is omitted for better readability.}
    \label{fig:pinn-settler}
\end{figure}

To determine the architecture hyperparameters of the PINN, we utilize a grid search varying the following hyperparameters: number of the hidden layers $\in \{1,2,3,4\}$ and width of hidden layers (number of nodes) $\in \{16, 32, 64, 128\}$. We find that a network with 2 hidden layers and 32 nodes performs the best on the validation trajectory. The parameters of the PINN models ($\bm{\theta}$) are initialized with the Xavier normal distribution (\cite{pmlr-v9-glorot10a}). All input variables are normalized to lie within the range $[-1,1]$ prior to training. After each layer except for the output layer, we use the tanh activation function. We use the sigmoid activation function for the output layer to prevent the argument of the square root in the denominator of Equations \eqref{eq:settler-2} and \eqref{eq:settler-3} from attaining negative values during PINN training.  The architecture of the PINN-based dynamic model is shown in \cref{fig:pinn-settler}.

\subsection{Physics equations and modeling assumptions}
\label{sec:physics_residuals}

The physics-informed component of the proposed PINN is based on the volume-balance equations governing the phase height dynamics in the gravity settler, as introduced in \cref{sec:mm_lls}. In the present work, only these macroscopic balance equations are incorporated into the PINN as soft constraints, while the detailed droplet coalescence and sedimentation submodels are not embedded.

Based on the volume balance equations in \cref{eq:settler}, the physics constraints are formulated as residuals that penalize deviations from the underlying dynamics. Specifically, we define an algebraic residual ($e_1$) describing the overall flow balance of the separator, and differential residuals ($e_2$ and $e_3$) describing the phase height dynamics:

\begin{subequations}\label{eq:residuals}
\begin{align}
    e_1(t) &= g_{\mathrm{sep}}\big(Q_{\mathrm{in}}(t), \hat{Q}_{\mathrm{bot}}(t), \hat{Q}_{\mathrm{top}}(t)\big), \\
    e_2(t) &= \dot{\hat{h}}_{\mathrm{DP}}(t) - f_{\mathrm{DP}}\big(\hat{h}_{\mathrm{DP}}(t), \hat{h}_{\mathrm{HP}}(t), Q_{\mathrm{in}}(t), \hat{Q}_{\mathrm{bot}}(t), \hat{Q}_\mathrm{c}(t), \hat{Q}_\mathrm{s}(t)\big), \\
    e_3(t) &= \dot{\hat{h}}_{\mathrm{HP}}(t) - f_{\mathrm{HP}}\big(\hat{h}_{\mathrm{HP}}(t), Q_{\mathrm{in}}(t), \hat{Q}_{\mathrm{bot}}(t), \hat{Q}_\mathrm{c}(t), \hat{Q}_\mathrm{s}(t)\big)
\end{align}
\end{subequations}

Here, $f_{\mathrm{DP}}$ and $f_{\mathrm{HP}}$ denote the right-hand sides of the volume-balance equations defined in \cref{sec:mm_lls}. The time derivatives of the predicted phase heights, $\dot{\hat{h}}$, are obtained via automatic differentiation of the neural network outputs with respect to the input time. These residuals define the physics-informed component of the training objective and penalize deviations from the governing equations.

Restricting the PINN to the volume-balance equations is mainly motivated by computational considerations. The semi-empirical submodels that define $Q_{\mathrm{c}}(t)$ and $Q_{\mathrm{s}}(t)$ as functions of the system states (cf. \cref{eq:submodels}) are lumped formulations obtained by resolving discretized internal compartments and subsequently aggregating their contributions. In particular, the separator is discretized along the axial direction into 200 segments and across 50 droplet diameter classes, resulting in approximately 200 × 50 = 10,000 internal compartments. Embedding these calculations into the PINN would either require introducing a tremendous number of additional outputs to represent the states of the discretized compartments or evaluating the submodels sequentially using PINN-predicted states during training. In both cases, this would lead to a prohibitive computational cost, as the full set of discretized compartments would need to be evaluated repeatedly at each training iteration. Moreover, the internal states associated with these submodels are not measurable in the experimental setup. 

Instead of embedding the submodels for $Q_\mathrm{c}(t)$ and $Q_\mathrm{s}(t)$ into the PINN, the influence of these submodels is incorporated indirectly through the synthetic data used during pretraining, which is generated using the full mechanistic model. In this way, the PINN may learn an implicit representation of the underlying physics while maintaining a tractable model structure. 

This modeling choice introduces a trade-off between physical completeness and computational efficiency. While the reduced physics representation may limit extrapolation in regimes dominated by droplet-scale phenomena, the combination of physics-informed training and experimental fine-tuning enables accurate phase height estimation in the operating conditions considered, as will be shown in \cref{sec:results_discussion}.

\subsection{Data handling for PINN training}
\label{sec:data_handling}

Based on the preprocessed experimental dataset $\mathcal{D}_{\mathrm{exp}}$ described in \cref{sec:exp-postproc}, phase height and flow-rate measurements are available at a uniform sampling interval of \SI{1}{\second}. This results in 789 segments (i.e., training samples), each corresponding to a single-step transition over \SI{1}{\second}. To align with the PINN formulation, which is defined over a temporal domain of $t \in [0,1]\,\si{\second}$, the trajectories are partitioned into consecutive segments of duration \SI{1}{\second}. Each segment represents a single-step state transition from process time $\tau_k$ to $\tau_{k+1} = \tau_k + \SI{1}{\second}$ and contains two measurement points for the phase heights, corresponding to the segment boundaries at $t=0$ and $t=1$.

For each segment, the PINN takes as input the initial states, the applied control input, and a specified value of the continuous time variable $t \in [0,1]$, and returns the corresponding system states and outputs at that time. The explicit inclusion of time as an input enables the use of automatic differentiation to evaluate time derivatives of the network outputs. At the same time, this formulation provides a continuous-time representation of the system dynamics within each segment, allowing the PINN to generate predictions at arbitrary intermediate times.

An analogous procedure is applied to the simulation dataset $\mathcal{D}_{\mathrm{sim}}$ generated from the mechanistic model described in \cref{sec:mm_lls}. Specifically, we generate 1000 segments of duration \SI{1}{\second}, with inputs sampled within the bounds of the experimental data (cf.~\cref{tab:bounds}). Within each segment, the mechanistic model is evaluated at a finer temporal resolution of $\Delta t = \SI{0.1}{\second}$, resulting in additional training points that enhance the representation of the system dynamics over the interval.

In addition to simulation and experimental datasets, $\mathcal{D}_{\mathrm{sim}}$ and $\mathcal{D}_{\mathrm{exp}}$, respectively, we create two more datasets, $\mathcal{D}_{\mathrm{physics}}$ and $\mathcal{D}_{\mathrm{init}}$ for physics-based training and initial condition training, respectively. For the physics-based training, we sample $|\mathcal{D}_{\mathrm{physics}}| = 10000$ collocation points corresponding to the PINN inputs: (i) time $t$, (ii) control $\Qin$, (iii) initial states $\hp(0)$ and $\hw(0)$. We use the extrapolation minimum and maximum bounds for controls and initial states given in \cref{tab:bounds}, and sample using LHS \citep{Iman1981AnAssessment}. Similarly, we choose $|\mathcal{D}_{\mathrm{init}}| = 1000$ collocation points as initial conditions within the same bounds at $t=0$. We use $\mathcal{D}_\mathrm{physics}$ and the initial condition dataset $\mathcal{D}_\mathrm{init}$ in both training stages (pretraining and fine-tuning). Note that, for stable training, we scale all the heights and the flow rates by division by the separator height $h_\mathrm{sep} = \SI{0.2}{\meter}$ and by a prescribed flow rate $Q_\mathrm{scale} = \SI{0.001}{\cubic\meter\per\second}$ respectively. 

It is important to note that, while the PINN yields continuous-time predictions, the experimental data provide measurements only at the discrete sampling instants corresponding to the segment boundaries. Intermediate predictions within each segment are therefore not directly supervised by experimental data, but are supported during training by both the pretraining dataset $\mathcal{D}_{\mathrm{sim}}$, which provides more densely sampled data, and the collocation points in $\mathcal{D}_{\mathrm{physics}}$. The validation and test trajectories are not included in $\mathcal{D}_{\mathrm{exp}}$ and are used exclusively for model evaluation.

\subsection{Training strategy}\label{sec:training}

Due to the considerable noise present in both the training and testing experimental trajectories, as well as the limited amount of data available, we use a two-stage training process, similar to \citet{Chakraborty2021}. First, the PINN is pretrained with the synthetic dataset $\mathcal{D}_\mathrm{sim}$ generated by the mechanistic model in \cref{sec:mm_lls}. The pretraining stage establishes baseline network weights and aligns the PINN with the dynamics of the mechanistic model before scarce and noisy experimental data is introduced. 

\begin{figure}
    \centering
    \includegraphics[scale=0.5]{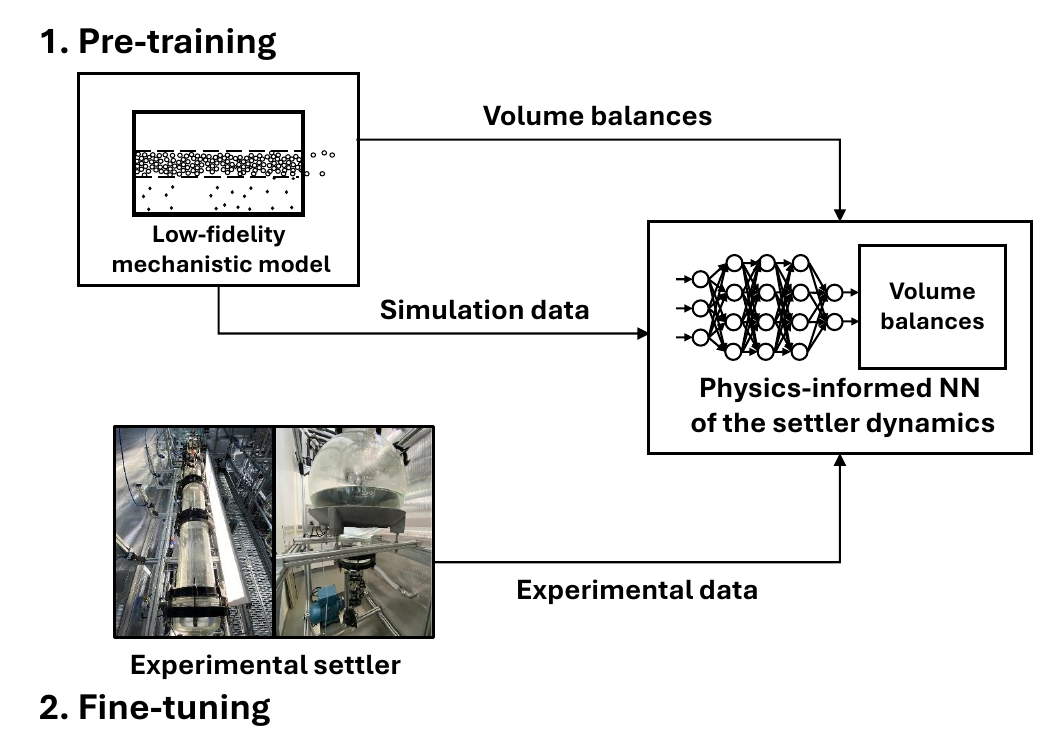}
    \caption{Two-stage training strategy of PINN-based liquid-liquid separator model}
    \label{fig:training-strategy}
\end{figure}

In the second stage, we fine-tune the pretrained models using data from the experimental setup, $\mathcal{D}_\mathrm{exp}$. The experimental dataset used for fine-tuning includes measurements of both phase heights and flow rates. To prevent the model from forgetting what it has learned during pretraining, we reduce the learning rate and the number of training epochs. The overall training strategy is depicted in \cref{fig:training-strategy}.

Note that the immeasurable internal flows $\Vz(t) = [Q_{\mathrm{c}}(t), Q_{\mathrm{s}}(t)]^\top$ are outputs of the mechanistic model. We therefore use these quantities as training data during the pretraining stage. During fine-tuning, however, their corresponding loss terms are disabled, as these internal flows are not measurable and thus no experimental data are available for them. 

The PINN parameters $\mathbf{\theta}$ can be learned by minimizing the mean squared error (MSE) loss, similar to \citet{Raissi2019Physics-informedEquations} and \citet{VELIOGLU2025108899}:
\begin{subequations}
    \allowdisplaybreaks
    \label{eq:pinn-loss-ic-cv}
    \begin{align}
    MSE_{total} &=  \lambda_\mathrm{data} MSE_\mathrm{data} + \lambda_\mathrm{latent} MSE_\mathrm{latent} + \lambda_\mathrm{init} MSE_\mathrm{init} \\ &+ \lambda_\mathrm{diff} MSE_\mathrm{diff} +  \lambda_\mathrm{alg} MSE_\mathrm{alg}, \label{eq:totalloss} \\
    \begin{split} \label{eq:data-loss}MSE_\mathrm{data} &= \frac{1}{4 |\mathcal{D}_{\mathrm{data}}|} \sum_{j=1}^{|\mathcal{D}_{\mathrm{data}}|} \Biggl[\left(\hpt(t_j) -  \hp(t_j)\right)^2 + \left(\hwt(t_j) -  \hw(t_j)\right)^2 \\
    &+ \left(\Qbt(t_j) - \Qb(t_j)\right)^2 + \left(\Qtt(t_j) -  \Qt(t_j)\right)^2\Biggr] \end{split}\\
    MSE_\mathrm{latent} &= \frac{1}{2 |\mathcal{D}_{\mathrm{data}}|} \sum_{j=1}^{|\mathcal{D}_{\mathrm{latent}}|}\Biggl[\left(\hat{Q}_\mathrm{c}(t_j) - Q_\mathrm{c}(t_j)\right)^2 + \left(\hat{Q}_\mathrm{s}(t_j) - Q_\mathrm{s}(t_j)\right)^2\Biggr]\\
    \label{eq:init-loss}MSE_\mathrm{init} &= \frac{1}{2 |\mathcal{D}_{\mathrm{init}}|}  \sum_{j=1}^{|\mathcal{D}_{\mathrm{init}}|} \Biggl[\left(\hat{h}_{\textrm{DP}_j}(t=0) - h_{\textrm{DP}_j}(0)\right)^2 - \left(\hat{h}_{\textrm{HP}_j}(t=0) - h_{\textrm{HP}_j}(0)\right)^2 \Biggr]\\
    \label{eq:res-loss}MSE_\mathrm{diff}  &= \frac{1}{2 |\mathcal{D}_{\mathrm{physics}}|} \sum_{j=1}^{|\mathcal{D}_{\mathrm{physics}}|} \Biggl[\left(e_2(t_j)\right)^2 
    + \left(e_3(t_j)\right)^2\Biggr]\\
    \label{eq:alg-loss}MSE_\mathrm{algebraic}  &= \frac{1}{|\mathcal{D}_{\mathrm{physics}}|} \sum_{j=1}^{|\mathcal{D}_{\mathrm{physics}}|} (e_1(t_j))^2
    \end{align}
\end{subequations}  

Here, $MSE_{\mathrm{data}}$ denotes the loss term associated with measurement data for phase heights and outlet flow rates, while $MSE_{\mathrm{latent}}$ corresponds to the loss for the latent internal flow rates. $MSE_{\mathrm{init}}$ quantifies the mismatch between the neural network predictions at $t = 0$ and the prescribed initial values $h_{\mathrm{HP}_j}(0)$ and $h_{\mathrm{DP}_j}(0)$. The terms $MSE_{\mathrm{diff}}$ and $MSE_{\mathrm{alg}}$ represent the physics-based loss contributions derived from the differential and algebraic residuals defined in \cref{eq:residuals}. The subscript $j$ indexes a finite set of samples taken at times $t_j$, each associated with initial values $h_{\textrm{HP}_j}(0)$ and $h_{\textrm{DP}_j}(0)$ and control action $Q_{in,j}$; for simplicity, initial values and control action are omitted from the notation. $\lambda_{\mathrm{data}}$, $\lambda_{\mathrm{latent}}$, $\lambda_{\mathrm{diff}}$, $\lambda_{\mathrm{alg}}$, and $\lambda_{\mathrm{init}}$ denote the weights assigned to the data, latent, differential physics, algebraic physics, and initial condition loss terms, respectively. $\mathcal{D}_\mathrm{data}$ refers to the dataset consisting of measurements; specifically, we use $\mathcal{D}_\mathrm{sim}$ in the pretraining stage and $\mathcal{D}_\mathrm{exp}$ in the fine-tuning stage. The simulation dataset $\mathcal{D}_{\mathrm{sim}}$ contains the internal flows $Q_{\mathrm{c}}(t)$ and $Q_{\mathrm{s}}(t)$. These quantities are absent from the experimental dataset $\mathcal{D}_{\mathrm{exp}}$; consequently, fine-tuning is performed with the corresponding loss weight for the latent internal flows set to zero ($\lambda_\mathrm{latent}$ = 0) . Note that the samples used to compute $MSE_{\mathrm{data}}$ and $MSE_{\mathrm{latent}}$ are drawn from the same dataset $\mathcal{D}_{\mathrm{data}}$, and similarly the samples used for $MSE_{\mathrm{diff}}$ and $MSE_{\mathrm{alg}}$ are both drawn from $\mathcal{D}_{\mathrm{physics}}$.

\begin{figure}
    \centering
    \begin{subfigure}{.47\textwidth}
        \centering
        \includegraphics[width=\linewidth]{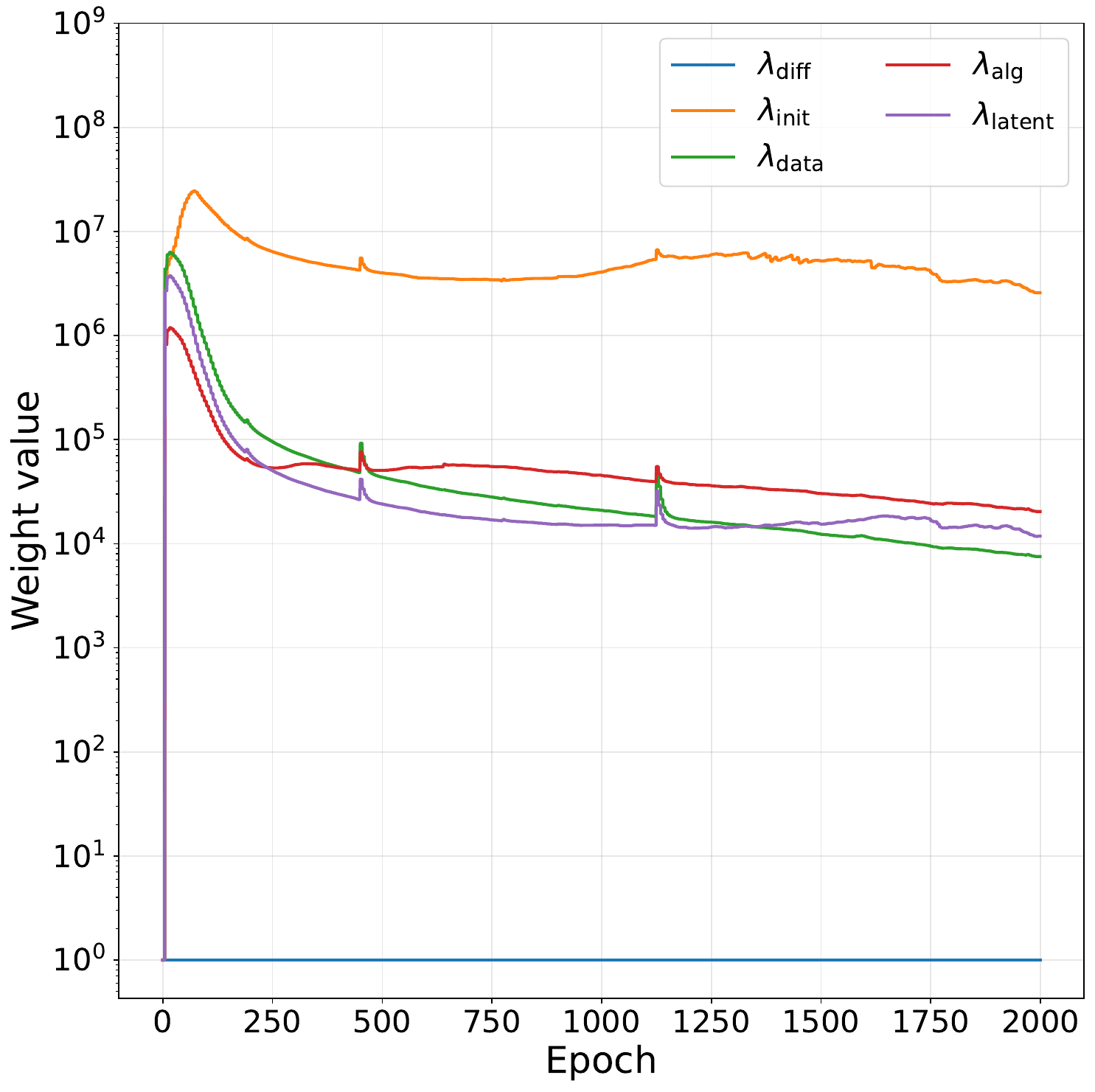}
        \caption{\textbf{IDW weights}}
        \label{fig:weights}
    \end{subfigure}%
    \begin{subfigure}{.47\textwidth}
        \centering
        \includegraphics[width=\linewidth]{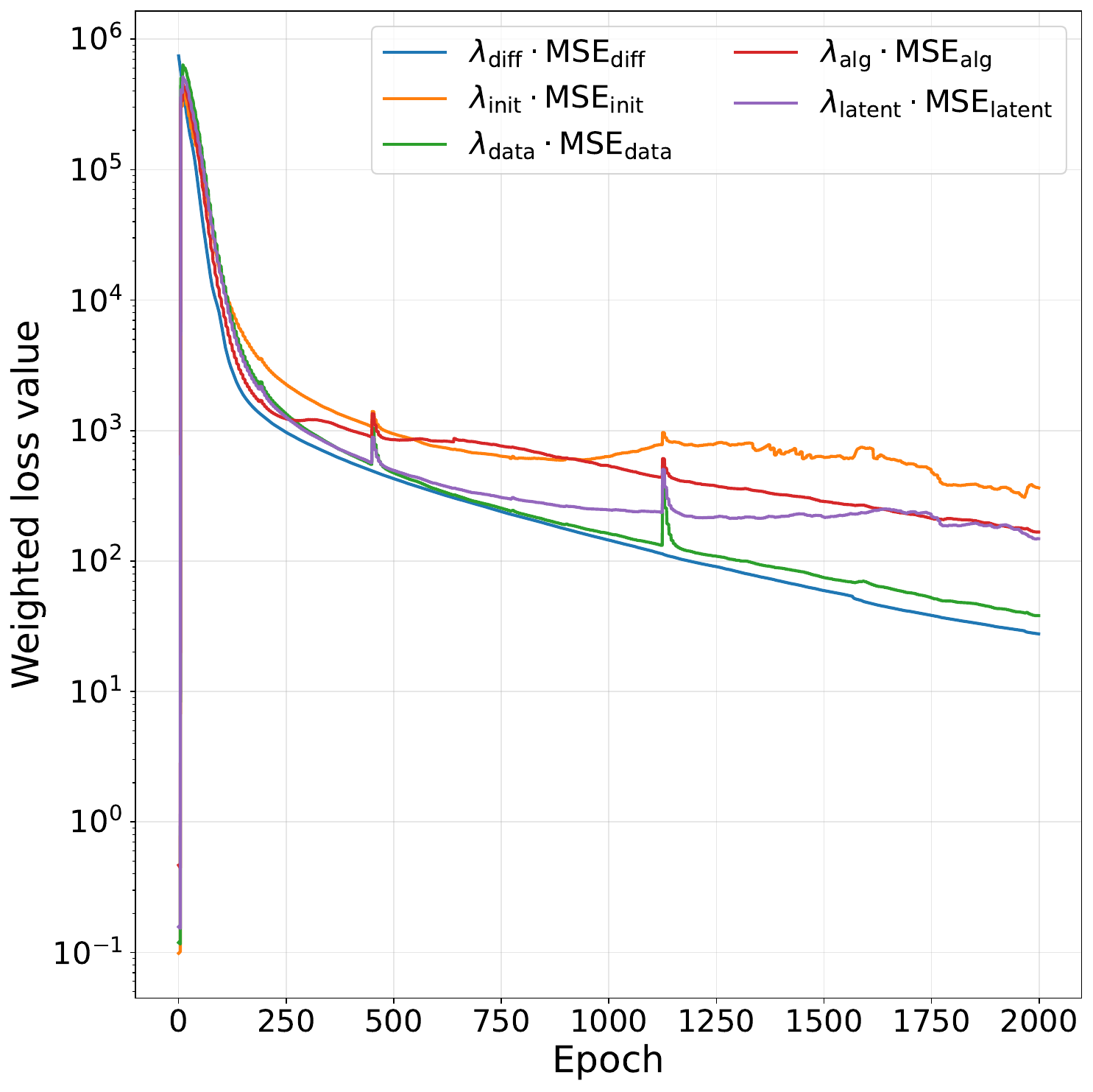}
        \caption{\textbf{Weighted loss terms}}
        \label{fig:losses}
    \end{subfigure}
    \caption{Evolution of adaptive loss weights (IDW) and corresponding weighted loss terms during pre-training, averaged over the ensemble. During fine-tuning, the weights are fixed to the final values from the pre-training.}
    \label{fig:idw}
    \end{figure}

We employ a two-optimizer training scheme for the PINN in each training stage. We start the training with a stochastic gradient descent (SGD) based optimizer (ADAM) \citep{kingma2017adammethodstochasticoptimization}, then switch to Limited-memory Broyden–Fletcher–Goldfarb–Shanno algorithm (L-BFGS) \citep{liu1989limited} for refinement. L-BFGS has been shown to be a very successful optimizer for training PINNs \citep{Markidis2021TheSolvers}, however, since it is a local optimizer, it has a tendency to get stuck in local minima \citep{Markidis2021TheSolvers}. \citet{du2019gradientdescentfindsglobal} have shown that SGD, due to its stochastic directions taken in optimization landscape can lead to (near) global training. Thus, we first use SGD to mitigate getting stuck in a local minimum, then we use L-BFGS to refine the network weights. In the pretraining stage, we use a fixed number of training epochs for each optimizer, 2000 for ADAM with a fixed learning rate of 0.001 and 300 for L-BFGS with the strong-Wolfe condition check activated. Moreover, we apply inverse Dirichlet weighting (IDW) \citep{Maddu2022InverseNetworks} every five epochs during the ADAM optimization stage to dynamically update the individual loss term weights, $\lambda_\mathrm{data}$, $\lambda_\mathrm{diff}$, $\lambda_\mathrm{init}$, $\lambda_\mathrm{alg}$ and $\lambda_\mathrm{latent}$ in Eq.~\eqref{eq:pinn-loss-ic-cv}. We note that IDW is adopted here as a practical training strategy based on preliminary empirical observations of improved stability and performance compared to a static weighting, consistent with its use in our previous work \citep{VELIOGLU2025108899}; a systematic quantitative evaluation of alternative weighting schemes is beyond the scope of this study. We report the evolution of the adaptive loss weights (IDW) and the corresponding weighted loss terms during pre-training in \cref{fig:idw}. Although the weights span several orders of magnitude, the weighted loss contributions are of comparable magnitude. In the fine-tuning stage, we reduce the learning rate by an order of magnitude to 0.0001 and lower the number of training epochs to 1000 for the ADAM optimizer and 200 for the L-BFGS optimizer. Moreover, we disable the IDW as the weightings of the different loss components are kept identical to the final values established in pretraining. 

\subsection{Benchmark vanilla neural network model} \label{sec:VNN}

As a comparison for the PINN, we implement a benchmark \emph{vanilla neural network} (VNN) model. The VNN has the same input configuration as the PINN: it takes time $t$, the current heights $\Vx(t=0) = [\hp(0),\hw(0)]^\top$ and control inputs $\Vu = [\Qin]$ as inputs and predicts the corresponding measurable heights $\hp(t)$, $\hw(t)$ and outlet flows $\Qb(t)$, $\Qt(t)$. Note that the VNN cannot predict the immeasurable internal flows $Q_{\mathrm{c}}(t)$ and $Q_{\mathrm{s}}(t)$, since no experimental data can be gathered for those. The VNN is trained in a data-driven fashion, i.e., without the physics-based loss terms in \cref{eq:res-loss} and \cref{eq:alg-loss}. For comparison with the PINN model, we adopt the same network architecture as used for the PINN in \cref{sec:training}, i.e, the same number of layers and neurons and the same activation functions. Moreover, we use the same datasets, except for, $\mathcal{D}_\mathrm{physics}$, which is used exclusively in physics-based training.

\subsection{Prediction results} \label{sec:validation}

We analyze how the models predict the experimental trajectories of phase heights when initialized from a \emph{known state} at $\tau=0$. We compare the accuracies of the PINN, VNN, and mechanistic model predictions, and examine the effects of pretraining. For both PINN and VNN, we employ an ensemble of models. Using an ensemble enables averaging over predictions to mitigate the impact of outlier models and provides a means to quantify prediction uncertainty \citep{dietterich2000ensemble}. The ensemble size for both PINN and VNN is fixed at $n_{\mathrm{ensemble}} = 40$. To construct the ensemble, we train $n_{\mathrm{ensemble}} = 40$ models independently with identical architectures and training data, but different random initialization of the network parameters. Due to the nonconvex loss landscape and stochasticity of the training algorithm, this may result in different trained parameter configurations and, consequently, a distribution of predictions. The ensemble spread is interpreted as an approximation of epistemic uncertainty arising from limited data and training variability.

\begin{figure}
\centering
\begin{subfigure}{.47\textwidth}
    \centering
    \includegraphics[width=\linewidth]{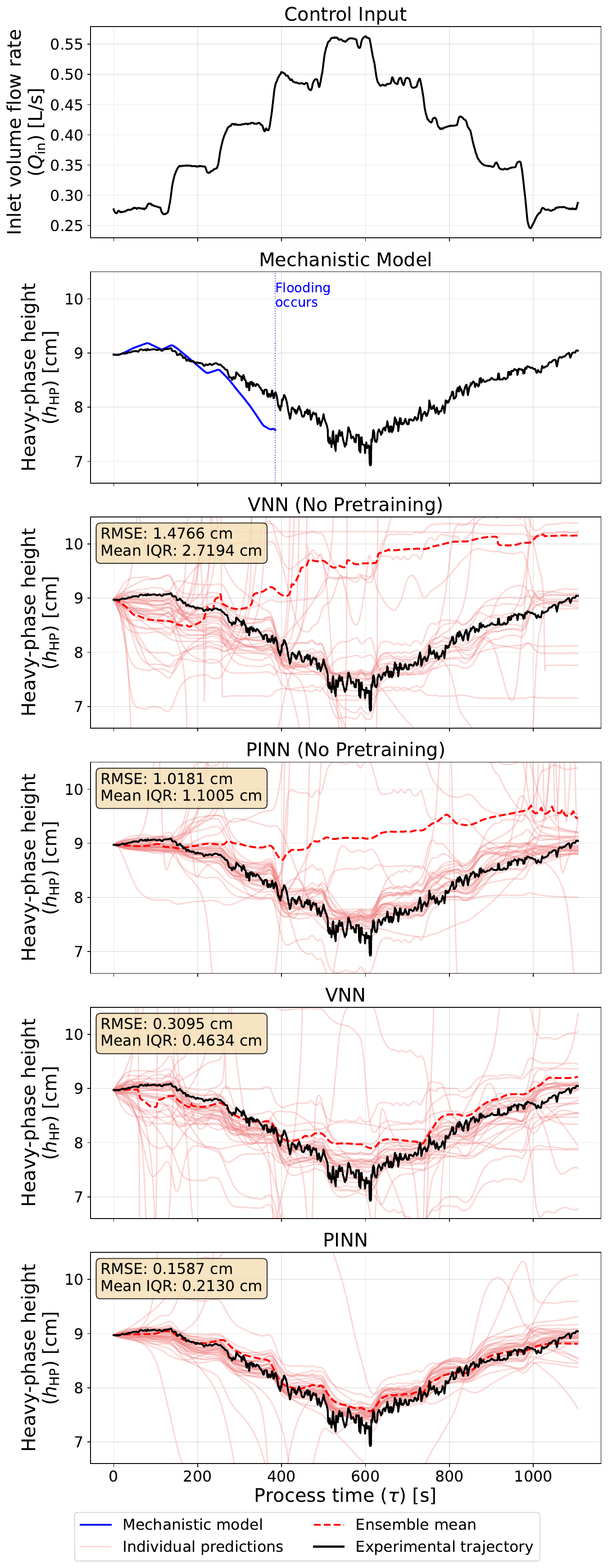}
    \caption{\textbf{Heavy-phase} (water) height}
    \label{fig:tr3-chaining-water}
\end{subfigure}%
\begin{subfigure}{.47\textwidth}
    \centering
    \includegraphics[width=\linewidth]{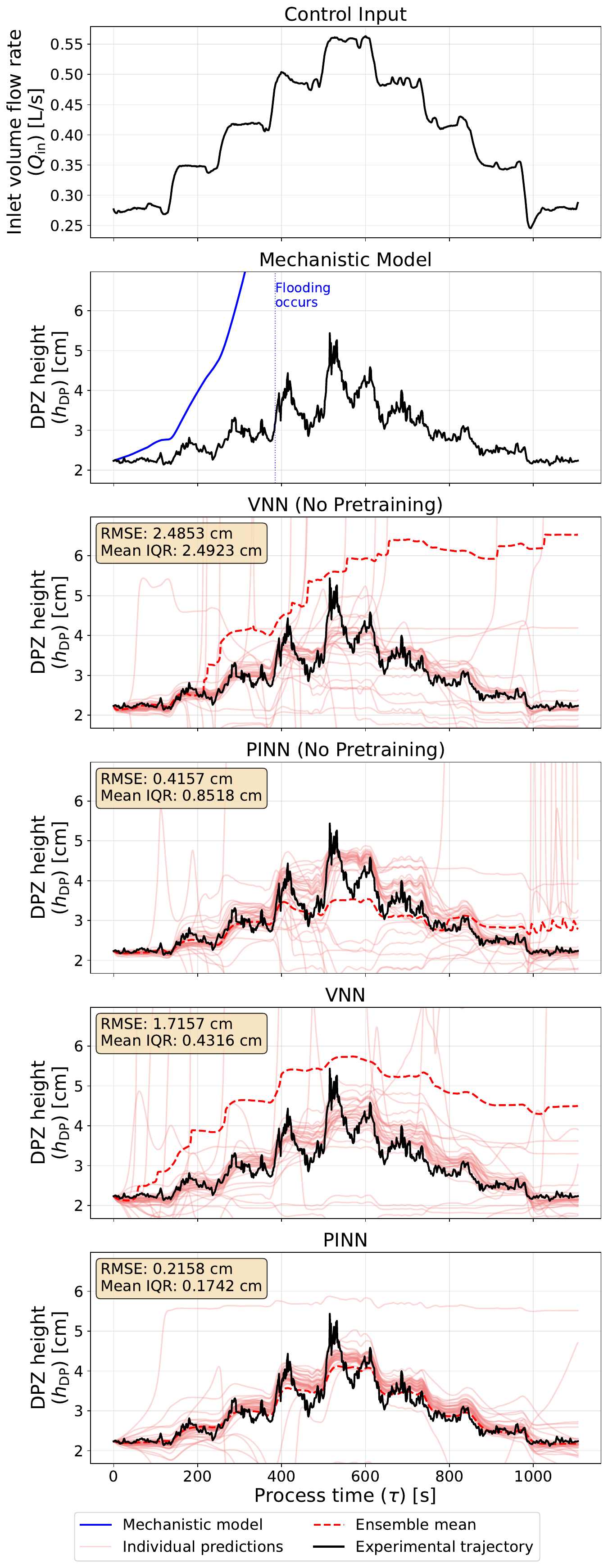}
    \caption{\textbf{Dense-packed zone} height}
    \label{fig:tr3-chaining-dpz}
\end{subfigure}
\caption{Simulation results for the \textbf{interpolation test trajectory}. Predictions from an individual model are shown in transparent red, whereas the ensemble mean estimation is shown with a red dashed line. The reported root mean square error (RMSE) values quantify the deviation between the ensemble mean prediction of each model and the corresponding experimental trajectory over the entire trajectory. Time-averaged IQR values are reported to quantify the ensemble spread.}
\label{fig:tr3-chaining}
\end{figure}

\begin{figure}
\centering
\begin{subfigure}{.46\textwidth}
    \centering
    \includegraphics[width=\linewidth]{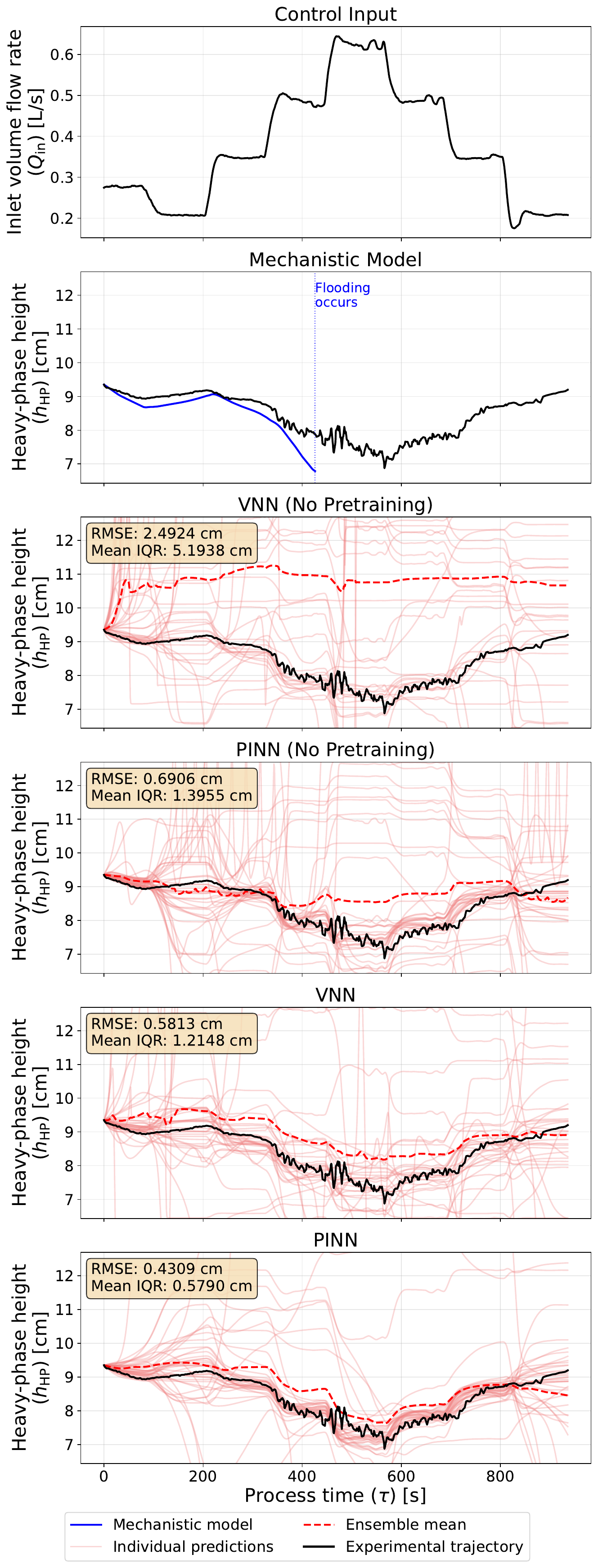}
    \caption{\textbf{Heavy-phase} (water) height}
    \label{fig:tr4-chaining-water}
\end{subfigure}%
\begin{subfigure}{.46\textwidth}
    \centering
    \includegraphics[width=\linewidth]{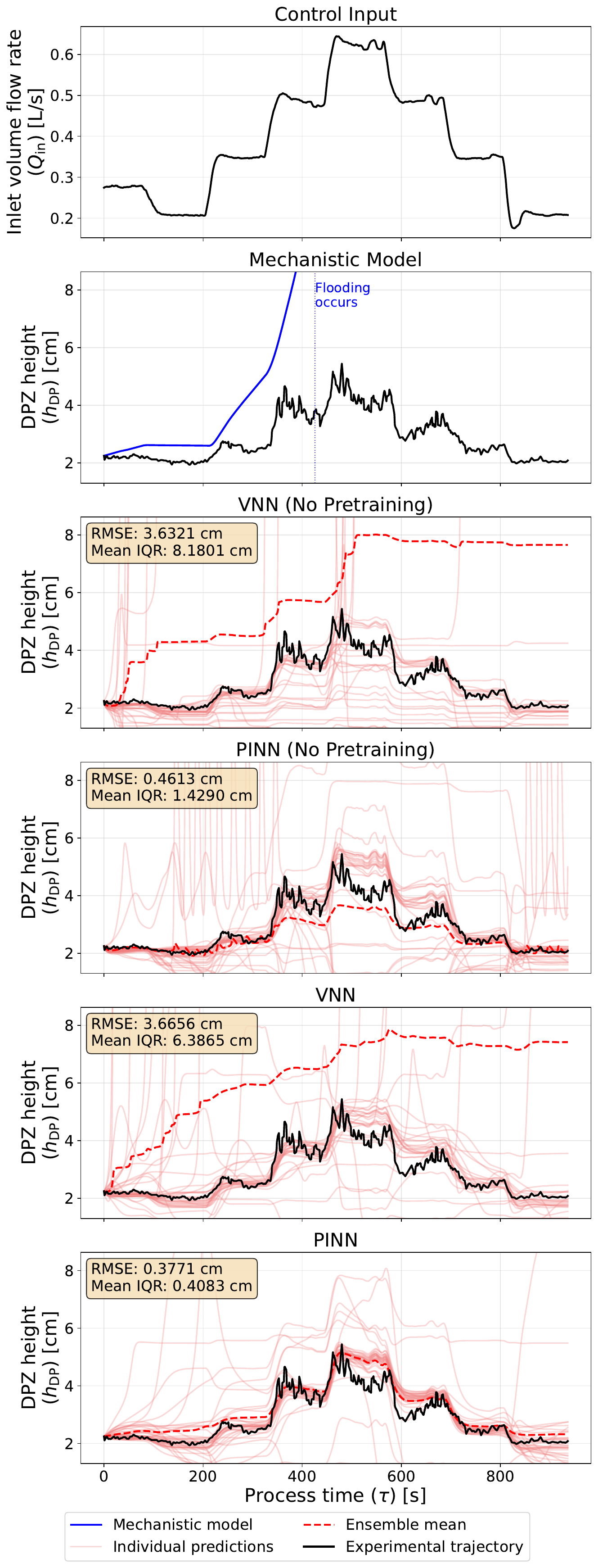}
    \caption{\textbf{Dense-packed zone} height}
    \label{fig:tr4-chaining-dpz}
\end{subfigure}
\caption{Simulation results for the \textbf{extrapolation test trajectory}. Predictions from an individual model are shown in transparent red, whereas the ensemble mean estimation is shown with a red dashed line. The reported RMSE values quantify the deviation between the ensemble mean prediction of each model and the corresponding experimental trajectory over the entire trajectory. Time-averaged IQR values are reported to quantify the ensemble spread.}
\label{fig:tr4-chaining}
\end{figure}

We report prediction results in \cref{fig:tr3-chaining} and \cref{fig:tr4-chaining}. To quantitatively assess model performance, we report the root mean square error (RMSE) of the ensemble-mean predictions over each trajectory, defined as
\[
\mathrm{RMSE} = \sqrt{\frac{1}{N} \sum_{i=1}^{N} \left( \hat{y}_i - y_i \right)^2},
\]
where $\hat{y}_i$ and $y_i$ denote the predicted and reference values, respectively, and $N$ is the number of samples. In addition, to quantify prediction uncertainty, we report the time-averaged interquartile range (IQR) \citep{dekking2005modern} of the ensemble predictions over each trajectory, defined as
\[
\mathrm{IQR} = q_{0.75} - q_{0.25},
\]
where $q_{0.25}$ and $q_{0.75}$ denote the 25th and 75th percentiles of the ensemble predictions at a given time instant, respectively. The IQR is used instead of the standard deviation to reduce sensitivity to a small number of outlier models \citep{dekking2005modern}.

Since the PINN and VNN models take time $t \in [0,1] \SI{}{\second}$, initial states, and control inputs as input and predict the heights and outlet flows at time $t$, trajectories over longer horizons must be constructed by chaining, i.e., the predicted heights is passed forward as the initial condition for the subsequent segment. Control inputs are known and provided for each segment of the trajectory. 

In Figs. \ref{fig:tr3-chaining-water} and \ref{fig:tr3-chaining-dpz}, we present the results for the interpolation trajectory for the heavy-phase height and the DPZ height, respectively. While the two-stage trained PINN ensemble tracks the averaged heights with good accuracy, the VNN ensemble mean overshoots with regard to the DPZ height. For both model types, the non-pretrained versions yield noticeably poorer predictions, with the effect being especially pronounced for the VNN. Moreover, we see that the mechanistic model predictions diverge, predicting flooding of the separator with the dispersed-phase at around $\tau = \SI{380}{\second}$. 

In Figs. \ref{fig:tr4-chaining-water} and \ref{fig:tr4-chaining-dpz}, we present the extrapolation trajectory results. The overall trends are similar to those observed for the interpolation trajectories: The two-stage trained PINN ensemble provides a fairly accurate tracking (albeit a slight overshoot towards the end), the VNN and the non-pretrained models perform significantly worse and the mechanistic model predicts flooding at around $ \tau = \SI{420}{\second}$.

These results demonstrate the effectiveness of the two-stage training procedure and physics-based regularization. Moreover, while neither the mechanistic model nor the PINNs solely trained on experimental data can produce high-quality predictions, the combination of the two (through pretraining and fine-tuning) enables the PINN ensemble to track the trajectories quite accurately. 

\section{Phase height estimation}\label{sec:results}

We now apply the models to estimate the phase heights in the gravity settler. It is important to distinguish between training and deployment: While phase height measurements are used during training, they will not be available during deployment. In \cref{sec:EKF} we present the state estimation framework that is inspired by the EKF to track and estimate the settler heights. Next, we describe the simple neural network to predict the DPZ height at separator outlet in \cref{sec:dpz_end}. Finally, in \cref{sec:results_discussion}, we discuss the estimation results, comparing PINN and VNN. 

\subsection{State estimation framework} \label{sec:EKF}
The EKF is the nonlinear version of the Kalman filter \citep{Kalman1960AProblems}, with the following state transition and measurement models in the discrete form \citep{Gelb:1974}:
\begin{align}
\Vx_k &= \bm f(\Vx_{k-1}, \Vu_{k-1}) + \mathbf{w}_{k-1}, \\
\Vy_k &= \bm h(\Vx_k) + \mathbf{v}_{k-1}
\end{align}
\noindent Here, subscript $k$ denotes the time step, $\bm f(\cdot)$ and $\bm h(\cdot)$ are the state-transition and measurement models, respectively, and $\mathbf{w}$ and $\mathbf{v}$ are process and measurement noise, respectively, that are assumed to be random Gaussian processes with zero mean and covariance $\mathbf{W}_k$ and $\mathbf{R}_k$ respectively \citep{Gelb:1974}. Note that, since $\bm f(\cdot)$ and $\bm h(\cdot)$ are nonlinear, they cannot be directly applied to the filter but have to be linearized around the current estimate, with Jacobians $\mathbf{F} = \frac{\partial \bm f}{\partial \mathbf{x}}\Big|_{\Vx_{k-1}}$ and $\mathbf{H} = \frac{\partial \bm h}{\partial \mathbf{x}}\Big|_{\Vx_{k}}$. The Jacobians of the neural networks (PINN and VNN) can be readily obtained by automatic differentiation \citep{naumannAD} of network outputs w.r.t. the network inputs \citep{Arnold2021StatespaceNetworks}.

We implement an EKF-inspired filter using the PINN as follows. We take subsets of the PINN outputs that are estimated and measured, i.e.,
\[
\Vx = [\hw,\hp]^\top, \: \mathbf{y} =[\Qb ,\Qt]^\top,
\]
respectively, where $\hw$ and $\hp$ are the states to be estimated by the filter, and $\Qb$ and $\Qt$ are the measurements. Note that the bottom outlet flow rate $\Qb$ is controlled via the valve position during the experimental campaign and therefore constitutes an input to the physical system, directly influencing the internal phase heights. Despite this causal direction, an inverse mapping from the phase heights to the outflows is learned by the PINN, which will constitute the measurement model of the filter. This does not imply that the outlet flows are treated as states of the physical system; rather, they are used as signals whose dependence on the internal states is captured implicitly by the PINN.

The PINN is trained on a time domain $t \in [0,1]\,\text{s}$, matching the process data sampling interval of length $\Delta\tau = 1\,\text{s}$. Let the PINN be defined as
\begin{equation*}
    \mathrm{PINN}_{\bm{\theta}}(\cdot):\;\mathbb{R}^{n_x} \times \mathbb{R}^{n_u} \times [0,1] \;\to\; \mathbb{R}^{n_x+n_y+n_z},
\end{equation*}
which maps the current estimated states $\Vx_{k-1}$, control input $\Vu$, and time $t \in [0,1]$ to a joint prediction vector of (next) estimated states $\Vx_k$, measurements $\Vy$ and immeasurable internal flows $\Vz$. We then partition (by notation) the PINN into two functions, i.e.,

\begin{equation*}
        \mathrm{PINN}_{\bm{\theta}}(\hat{\mathbf{x}}, \mathbf{u}, t) =
        \begin{bmatrix}
        \bm f_{\bm{\theta}}(\hat{\mathbf{x}}, \mathbf{u}, t) \\
        \bm h_{\bm{\theta}}(\hat{\mathbf{x}}, \mathbf{u}, t)
        \end{bmatrix},    
\end{equation*}

\noindent where $\bm f_{\bm{\theta}}(\cdot):\;\mathbb{R}^{n_{x}} \times \mathbb{R}^{n_u} \times [0,1] \;\to\; \mathbb{R}^{n_{x}}$ and $\bm h_{\bm{\theta}}(\cdot):\;\mathbb{R}^{n_{x}} \times \mathbb{R}^{n_u} \times [0,1] \;\to\; \mathbb{R}^{n_y}$ map the current estimated states, control input, and time $t \in [0,1]$ into predicted estimated states and predicted measurements, respectively. The outputs associated with the immeasurable internal flows $\Vz = [Q_{\mathrm{c}}(t), Q_{\mathrm{s}}(t)]^\top$ are omitted from the filter, since our PINN architecture does not allow computing Jacobians w.r.t. to these flows. We then define the \emph{state transition model} and \emph{the measurement model} by evaluating the PINN at $t=1$ and $t=0$ respectively:
\begin{align}
    \text{State transition model:}\quad & \hat{\mathbf{x}}_{k|k-1} = \bm f_{\bm{\theta}}(\hat{\mathbf{x}}_{k-1|k-1}, \mathbf{u}_{k-1}, t = 1) \\
    \text{Measurement model:}\quad & \hat{\mathbf{y}}_k = \bm h_{\bm{\theta}}(\hat{\mathbf{x}}_{k|k-1}, \mathbf{u}_{k-1}, t = 0) 
\end{align}
\noindent where the notation $\hat{\Vx}_{k|k-1}$ denotes the predicted state at time step $k$ based on information up to time step $k-1$. 

The prediction step \citep{Gelb:1974} is then used to obtain the predicted state and covariance estimates:
\begin{align}
\hat{\mathbf{x}}_{k|k-1} &= \bm f_{\bm{\theta}}(\hat{\mathbf{x}}_{k-1|k-1}, \mathbf{u}_{k-1}, t = 1) , \\
\mathbf{P}_{k|k-1} &= \mathbf{F}_{k} \mathbf{P}_{k-1|k-1} \mathbf{F}_{k}^\top + \mathbf{W}_k,
\end{align}
where $\mathbf{F}_k = \frac{\partial \bm f_{\bm{\theta}}}{\partial \mathbf{\Vx}}\big|_{\hat{\mathbf{x}}_{k-1|k-1}}$ is the Jacobian of the state transition model $\bm f_{\bm{\theta}}(\cdot)$ with respect to the estimated states, $\mathbf{P}_k$ is the state covariance matrix, and $\mathbf{W}_k$ is the process noise covariance at time step $k$. Here, we initialize  $\mathbf{P}_0$ as a diagonal matrix with values 0.0001 and 0.0001, corresponding to $\Vx = [\hw,\hp]^\top$. 

The process noise covariance $\mathbf{W}_k$ is computed adaptively at each time step. Such adaptive sampling of the covariance matrix $\mathbf{W}_k$, sometimes referred to as Adaptive Kalman Filtering \citep{mohamed1999adaptive}, is a well-established technique that has been shown to improve both the stability and performance of the Kalman Filter \citep{mohamed1999adaptive, zhang2020identification}. We leverage the model ensemble of 40 independently trained PINNs to calculate the process noise covariance $\mathbf{W}_k$ in the following way: (i) First, we take the ensemble mean of the last estimated state $\hat{\mathbf{x}}_{k-1|k-1}$ and feed it as the initial state to the transition model  $\bm f_{\bm{\theta}}(\cdot)$ of each ensemble member individually. (ii) We then obtain the next state estimate $\hat{\mathbf{x}}_{k|k-1}$ from each ensemble member. (iii) We calculate the sample covariance of these 40 predictions to obtain the process noise covariance $\mathbf{W}_k$ at each time step $k$. Here, the ensemble spread is interpreted as a sample-based approximation of the prediction uncertainty of the PINN model. The sample covariance of the ensemble predictions is used as a proxy for the process noise covariance $\mathbf{W}_k$, thereby incorporating model (epistemic) uncertainty into the filtering procedure. This allows the filter to account for increased uncertainty in regions where the model predictions diverge.

Next, we use the measurement model $\bm h_{\bm{\theta}}(\cdot)$ that takes the last estimated states as initial state and outputs corresponding (predicted) measurements:
\begin{equation}
\hat{\mathbf{y}}_{k} = \bm h_{\bm{\theta}}(\hat{\mathbf{x}}_{k|k-1}, \mathbf{u}_{k-1}, t = 0)
\end{equation}

The update equations are then used to update the state and covariance estimates using the residual of true and predicted measurements and the Kalman gain $\mathbf{K}_k$ \citep{Gelb:1974}:
\begin{align}
\mathbf{K}_k &= \mathbf{P}_{k|k-1} \mathbf{H}_k^\top \left( \mathbf{H}_k \mathbf{P}_{k|k-1} \mathbf{H}_k^\top + \mathbf{R}_k \right)^{-1}, \\
\hat{\mathbf{x}}_{k|k} &= \hat{\mathbf{x}}_{k|k-1} + \mathbf{K}_k\left( \mathbf{y}_k - \hat{\mathbf{y}}_{k} \right), \\
\label{eq:joseph}\mathbf{P}_{k|k} &= 
\bigl(\mathbf{I} - \mathbf{K}_k \mathbf{H}_k \bigr)\,
\mathbf{P}_{k|k-1}\,
\bigl(\mathbf{I} - \mathbf{K}_k \mathbf{H}_k \bigr)^{\!T}
+ \mathbf{K}_k \mathbf{R}_k \mathbf{K}_k^{T},
\end{align}
where $\mathbf{H}_k = \frac{\partial  \bm h_{\bm{\theta}}}{\partial \mathbf{x}}\big|_{\hat{\mathbf{x}}_{k|k-1}}$ is the Jacobian of the measurement model $\bm h_{\bm{\theta}}(\cdot)$ w.r.t to the estimated states. The notation $\hat{\Vx}_{k|k}$ represents the updated state at time step $k$ after incorporating the measurement at that step. The measurement noise covariance matrix $\mathbf{R}$ is diagonal and constant, with entries derived from the flow sensor specifications. Since both flow measurements come from the same sensor, a single $\mathbf{R}_k$ value is used for both states. The sensor specifies a maximum error of $e_{\mathrm{max}} = \pm 0.15\%$. Assuming a Gaussian error distribution, the standard deviation is approximated using the empirical $3\sigma$ heuristic as $\sigma = e_{\mathrm{max}}/3 = 5 \times 10^{-4}$ \citep{wackerly2008mathematical}. The resulting covariance is $\mathbf{R}_{\mathrm{sensor}} = \sigma^2 = 2.5 \times 10^{-7}$. Note that we use the Joseph form \citep{bucy2005filtering} for the covariance update in \cref{eq:joseph}, for improved stability. Preliminary analysis of the validation trajectory showed that the average ratio of $\frac{\mathbf{W}_k}{\mathbf{R}_k}$ was high, causing the filter to overtrust measurement updates and yield noisy state estimates. To address this, $\mathbf{W}_k$ was decreased by a factor of 100. After each update, the PINN initial state is reset to $\hat{\mathbf{x}}_{k|k}$ to ensure that the state transition model starts from the most recent state estimate. We apply the filter for each model in the ensemble separately, and we only use the ensemble mean of state estimates for the process noise covariance calculations.

\subsection{Prediction of DPZ height at the separator outlet} \label{sec:dpz_end}

In the wedge-shaped DPZ regime for higher flow rates, the maximum DPZ height consistently occurs at the end of the separator (cf. $h_{4,3}$ in \cref{fig:trajectory1_preprocessed}). In contrast, our models, assuming a band-shaped DPZ, predict the average DPZ height along the separator. Aiming at flooding detection, which requires monitoring the DPZ height close to the separator outlet, we construct a simple feedforward neural network (NN) that maps the average DPZ height to $h_{4,3}$. The network has two hidden layers with 16 and 8 nodes, and is trained in a purely data-driven fashion using data from the training trajectory. Training is performed with the Adam optimizer for up to 1000 epochs, with early stopping if the validation error (from the validation trajectory) does not improve after 30 epochs. After estimating the average phase heights with the PINN–EKF and VNN–EKF, we use the simple feedforward NN to predict $h_{4,3}$.

\subsection{Results} \label{sec:results_discussion}

\begin{figure}
\centering
\begin{subfigure}{.5\textwidth}
    \centering
    \includegraphics[width=\linewidth]{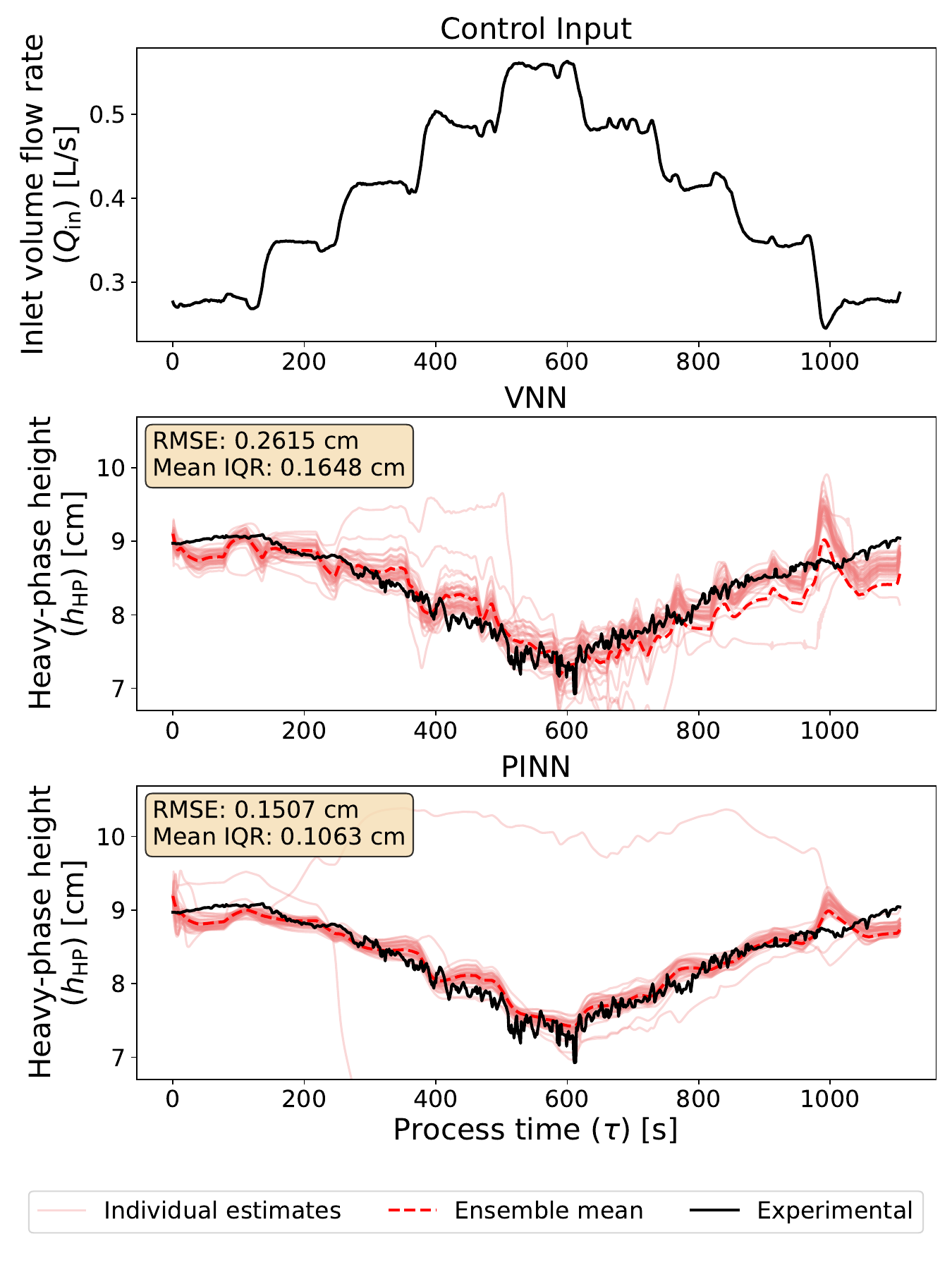}
    \caption{\textbf{Heavy-phase} (water) height}
    \label{fig:tr3-ekf-water}
\end{subfigure}%
\begin{subfigure}{.5\textwidth}
    \centering
    \includegraphics[width=\linewidth]{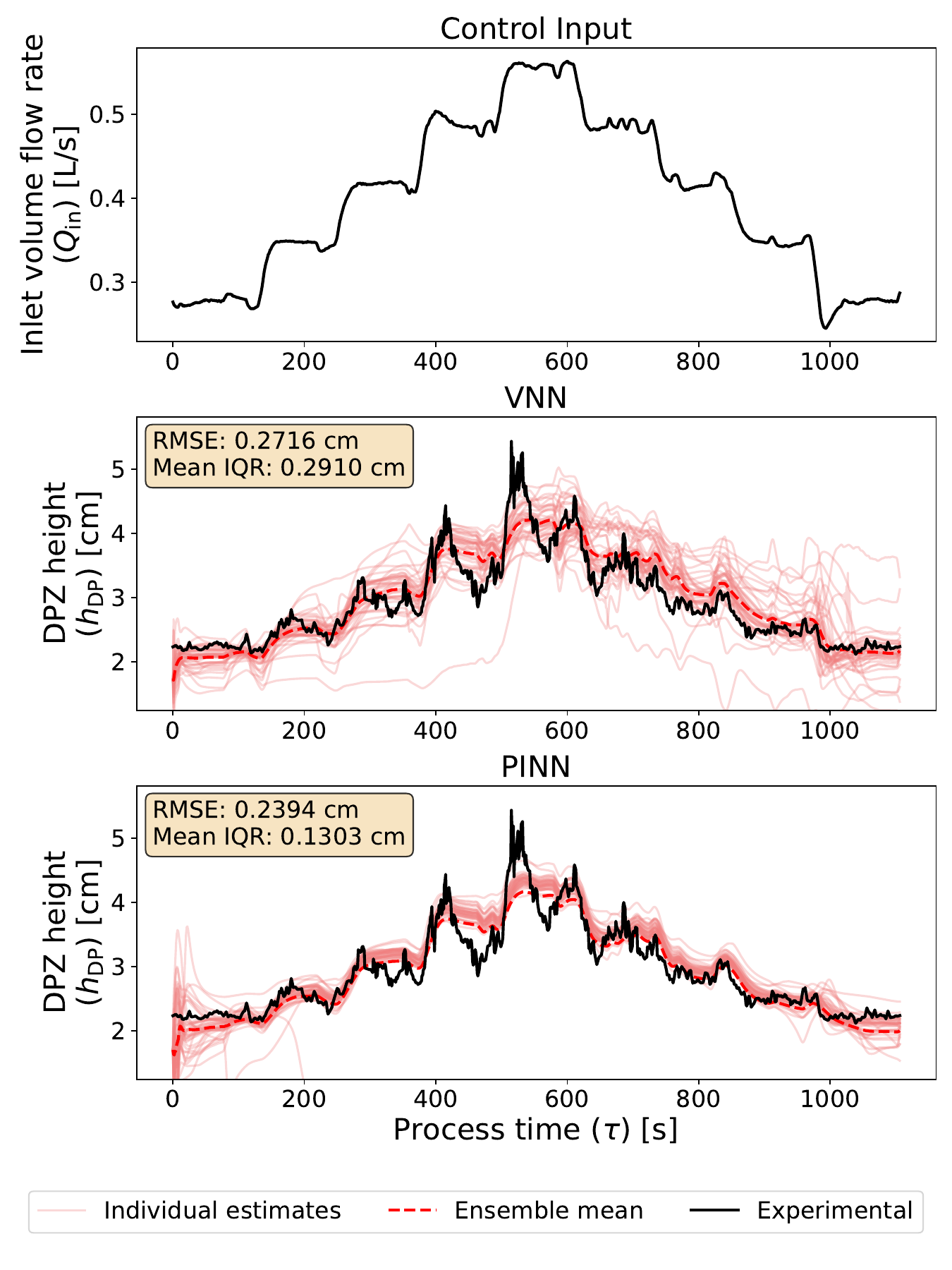}
    \caption{\textbf{Dense-packed zone} height}
    \label{fig:tr3-ekf-dpz}
\end{subfigure}
\caption{State estimation results for the \textbf{interpolation test trajectory}. Estimations from an individual model are shown in transparent red, whereas the ensemble mean estimation is shown with a red dashed line. The reported RMSE values quantify the deviation between the ensemble mean estimation of each model and the corresponding experimental trajectory over the entire trajectory. Time-averaged IQR values are reported to quantify the ensemble spread.}
\label{fig:tr3-EKF}
\end{figure}

\begin{figure}
\centering
\begin{subfigure}{.5\textwidth}
    \centering
    \includegraphics[width=\linewidth]{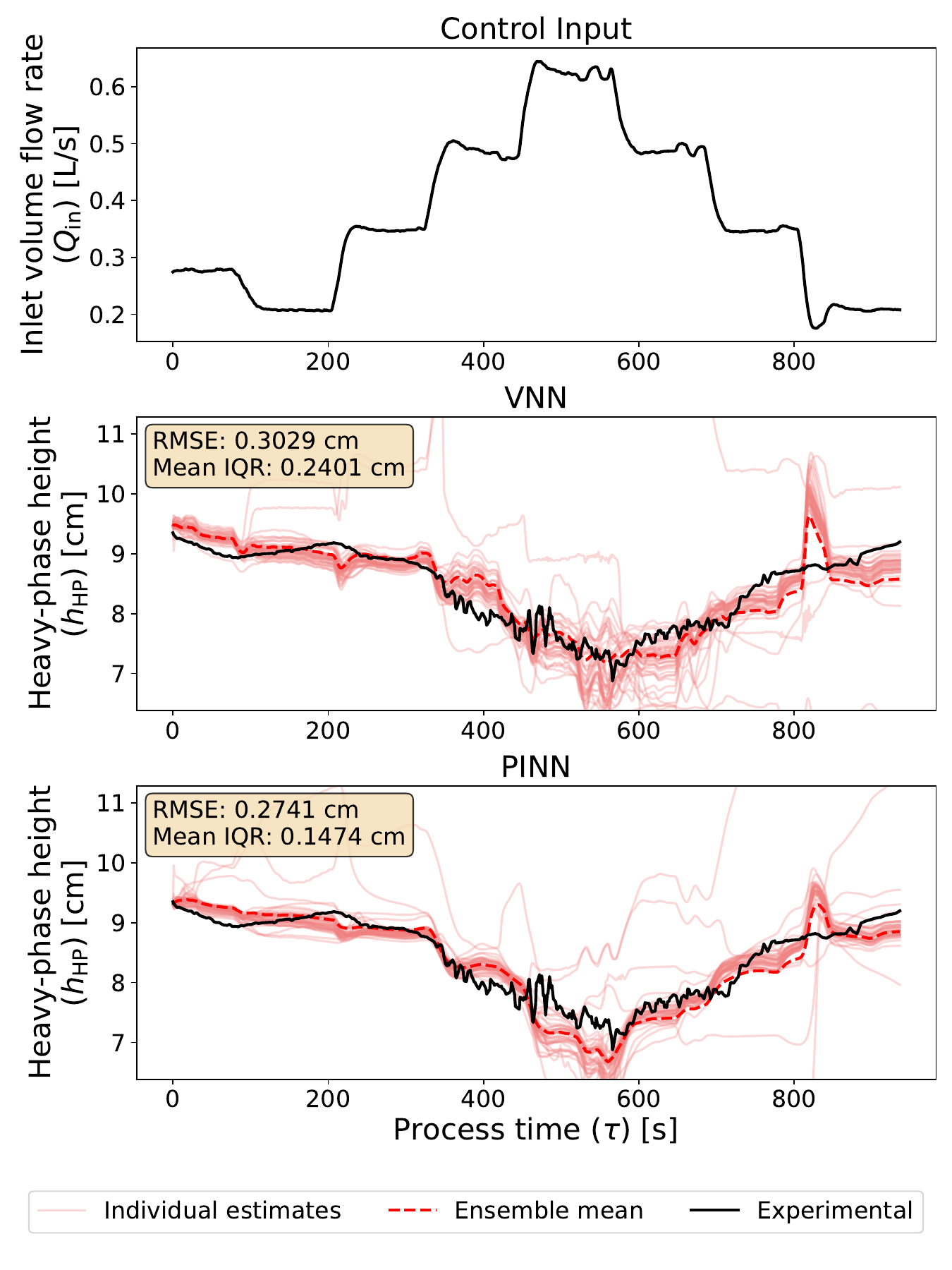}
    \caption{\textbf{Heavy-phase} (water) height}
    \label{fig:tr4-ekf-water}
\end{subfigure}%
\begin{subfigure}{.5\textwidth}
    \centering
    \includegraphics[width=\linewidth]{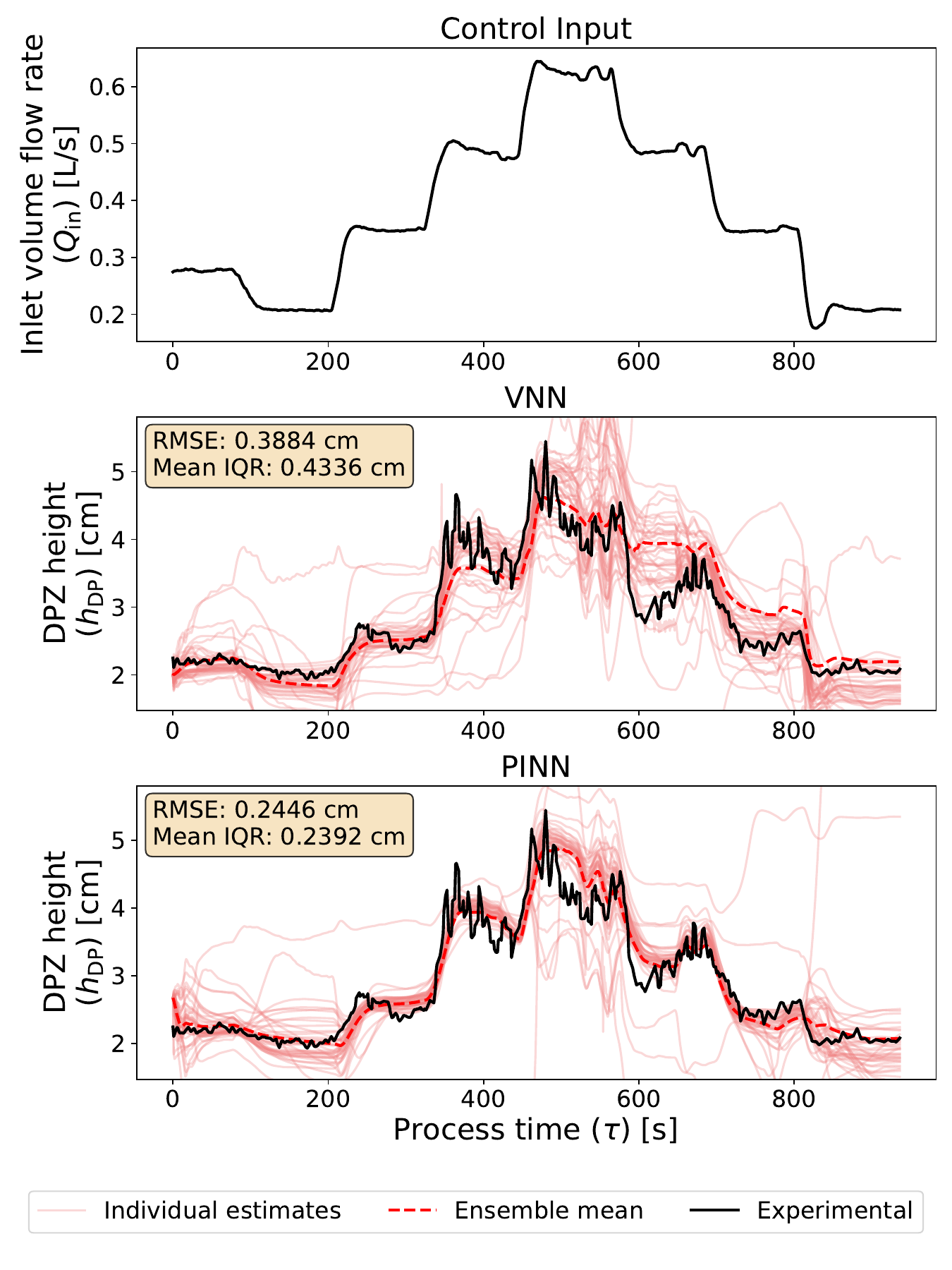}
    \caption{\textbf{Dense-packed zone} height}
    \label{fig:tr4-ekf-dpz}
\end{subfigure}
\caption{State estimation results for the \textbf{extrapolation test trajectory}. Estimations from an individual model are shown in transparent red, whereas the ensemble mean estimation is shown with a red dashed line. The reported RMSE values quantify the deviation between the ensemble mean estimation of each model and the corresponding experimental trajectory over the entire trajectory. Time-averaged IQR values are reported to quantify the ensemble spread.}
\label{fig:tr4-EKF}
\end{figure}

This section presents the results of the state estimation analysis. First, an initial guess for the phase heights is obtained by uniformly sampling 100 height vectors within their bounds and using those as initial state inputs to PINN (and VNN) to predict the outlet volume flows at $\tau = 0$. These predicted outflows are compared against the experimentally measured outlet flows at $\tau = 0$, and the candidate yielding the best agreement is selected as the initial state estimate.  We then feed these initial states into the filter and proceed to track and estimate the phase heights for each PINN (and VNN) model individually. The individual trajectories, along with the ensemble means, are shown in Figs. \ref{fig:tr3-EKF} and \ref{fig:tr4-EKF}. As in \cref{sec:results}, model performance and uncertainty are quantified using RMSE and time-averaged IQR, respectively.

In \cref{fig:tr3-EKF}, we report the results for the interpolation test trajectory. For both PINN and VNN, the estimated initial DPZ height $\hp(\tau_0)$ deviates from the experimental trajectory, indicating weak reconstruction of the internal states from outlet flow measurements alone. Nevertheless, in both cases the ensemble mean converges to the experimental trajectory at around $\tau = \SI{100}{\second}$. For the heavy phase height, the PINN consistently outperforms the VNN. Despite the significant overshoot observed in the forward simulation results shown in \cref{fig:tr3-chaining-dpz}, the VNN–filter combination achieves comparable accuracy to the PINN ensemble in predicting DPZ height for the interpolation trajectory. This indicates that the filter update step effectively compensates for the VNN’s error accumulation by incorporating outflow measurements. 
However, compared to the PINN, the individual estimates of the VNN ensemble exhibit lesser consistency, as shown by the lower IQR value of the PINN ensemble.

For the extrapolation test trajectory in \cref{fig:tr4-EKF}, the PINN ensemble performs better than the VNN ensemble in case of the DPZ height. In particular, after the maximum control action (an operating value unseen in the training trajectory) is applied at around process time $\tau = \SI{500}{\second}$, the VNN mean estimate overshoots the experimental DPZ height, whereas the PINN continues to track the DPZ height with good accuracy. Again, the individual estimates from the PINN ensemble are also more consistent than the ones from the VNN, as evidenced by the lower IQR value. For the heavy phase height, the VNN and PINN ensembles demonstrate similar accuracy, with the PINN underestimating the heavy phase height around process time $\tau = \SI{500}{\second}$, while the VNN shows increasing deviation from the experimental trajectory thereafter.

\begin{figure}
\centering
\begin{subfigure}{.5\textwidth}
    \centering
    \includegraphics[width=\linewidth]{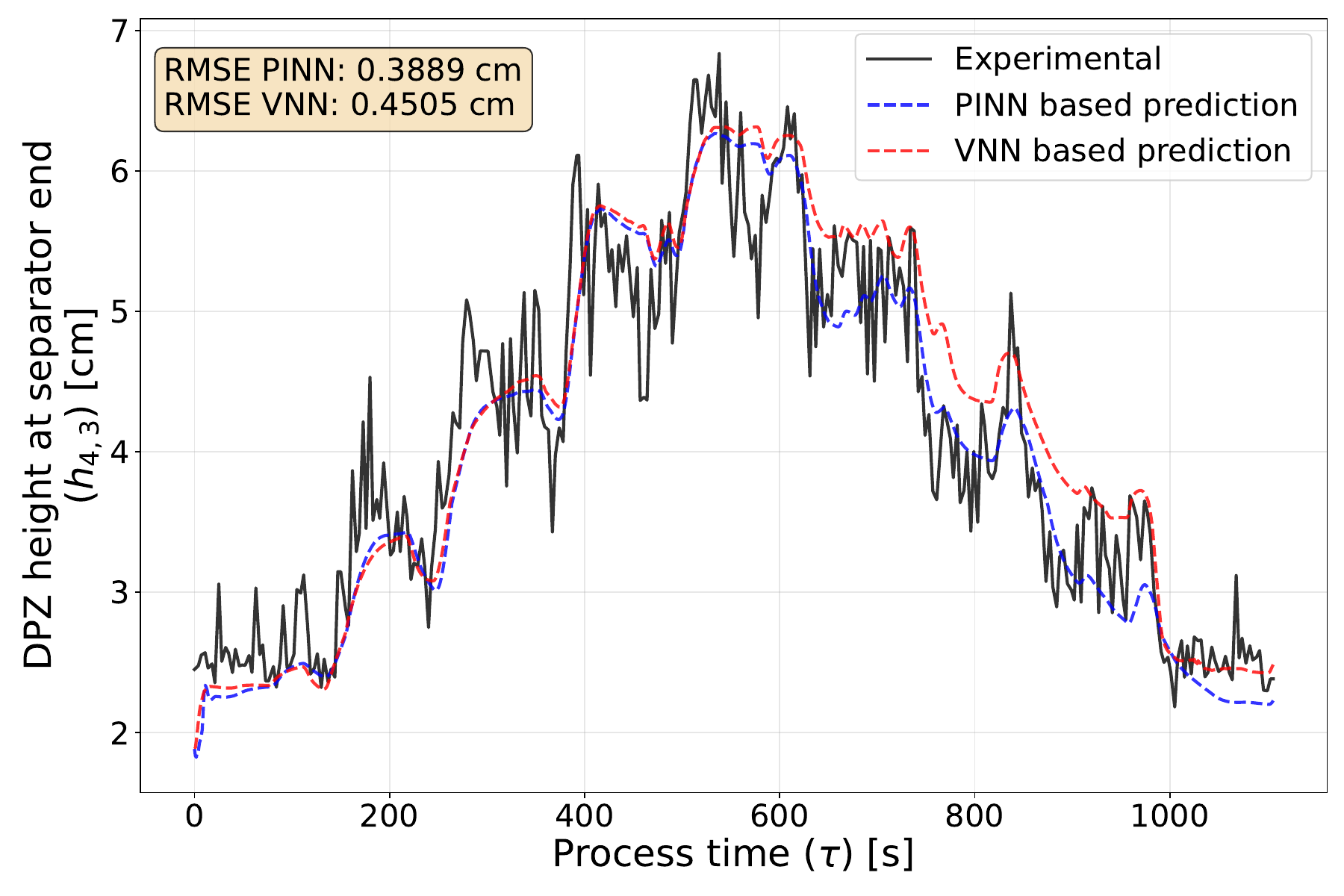}
    \caption{\textbf{Interpolation} test trajectory}
    \label{fig:tr3-dpz-end}
\end{subfigure}%
\begin{subfigure}{.5\textwidth}
    \centering
    \includegraphics[width=\linewidth]{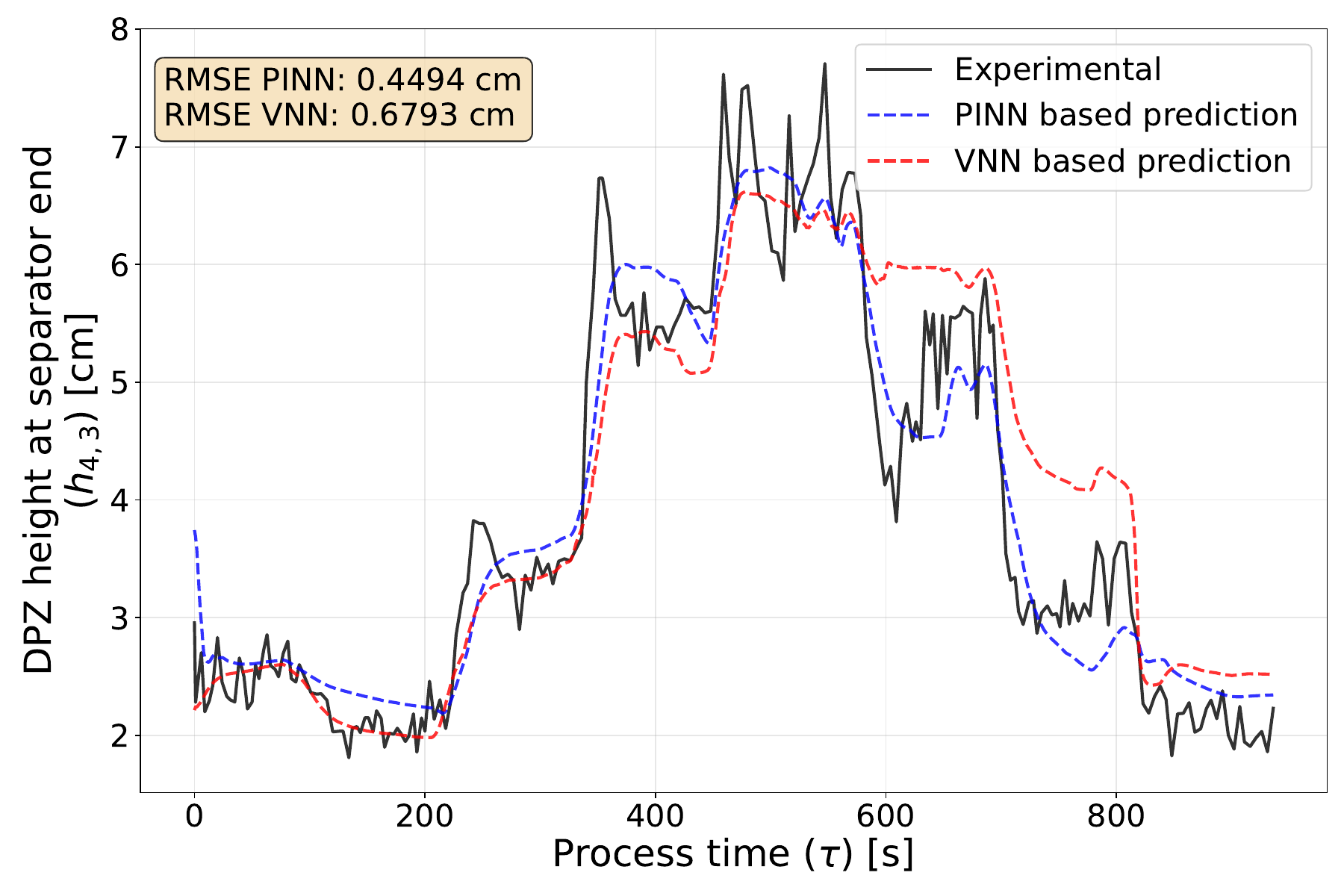}
    \caption{\textbf{Extrapolation} test trajectory}
    \label{fig:tr4-dpz-end}
\end{subfigure}
\caption{Prediction of the DPZ height at the separator end. Experimental results are obtained from camera \textbf{QIR04} at detection window $h_{4,3}$. Estimates of the average DPZ from both PINN and VNN are fed into the NN that predicts the DPZ height at separator end. The reported RMSE values quantify the deviation between the prediction of each model and the corresponding experimental trajectory over the entire trajectory.}
\label{fig:dpz-end}
\end{figure}

After estimating the average DPZ height along the separator using the filter, we apply the simple NN to predict the DPZ height at the separator end (\cref{fig:dpz-end}). Both the PINN and VNN ensembles demonstrate good accuracy. Deviations occur primarily when the average DPZ height itself deviates, indicating that the NN performs well, but errors propagate where the underlying ensemble predictions are inaccurate (cf. \cref{fig:tr3-EKF,fig:tr4-EKF}). This effect is emphasized in the extrapolation trajectory (cf. \cref{fig:tr4-EKF}), where larger deviations in the VNN ensemble lead to correspondingly larger errors in the predicted DPZ height at the separator end.

We therefore conclude that the two-stage trained PINN, when combined with the filter, enables effective real-time estimation of the internal separator states based on outlet flow measurements. As the outlet flow-rate measurements are inexpensive to obtain, our PINN constitutes a practical soft sensor for the difficult-to-measure phase heights. Moreover, we demonstrate that predicting the DPZ height at the separator end (the maximum value) from the average DPZ height is possible using a simple NN.

\section{Conclusion and Outlook}\label{sec:conclusion} 
We investigated the estimation of phase heights in a pilot-scale liquid–liquid separator operating with 1-octanol and deionized water at 30 °C under varying flow conditions. Direct measurement of phase heights is challenging, as the interior of the separator is not readily accessible for reliable observation of the phase interface. Although we employ a YOLOv8-based phase-boundary detection algorithm in our experimental setup, the resulting measurements still contain noise due to the difficulty of distinguishing the interface and the sensitivity of the algorithm to illumination conditions. Moreover, the available mechanistic model relies on multiple simplifying assumptions, resulting in limited predictive accuracy. To address these challenges, we developed a PINN-based dynamic model and proposed a two-stage training strategy: pretraining on approximate physics, i.e., synthetic data from the low-fidelity model, followed by fine-tuning with experimental data. We further used ensemble learning to mitigate the influence of outlier models. Moreover, the ensemble not only improves prediction robustness but also provides a practical measure of model uncertainty, which is incorporated into the state estimation framework through adaptive covariance estimation. Finally, we combined the trained PINN with an EKF-inspired filter to enable estimation of the aqueous phase and DPZ heights. The filter prevents error accumulation from chaining of PINN predictions by fusing model predictions with real-time flow measurements to update the estimated phase-height trajectories.

The two-stage PINN ensemble outperformed both the low-fidelity mechanistic model and a single-stage PINN ensemble trained only on experimental data, showing that integrating mechanistic model knowledge with experimental measurements yields better results than either approach alone. For further comparison, we also pretrained and fine-tuned an ensemble of purely data-driven vanilla neural networks (VNNs). Both the PINN and VNN ensembles were embedded in a state estimation framework to enable phase height estimation during deployment from flow-rate measurements only. The PINN ensemble achieved higher accuracy in tracking phase heights. Furthermore, a simple neural network can predict the outlet DPZ height (typically the maximum value and most critical for flooding) from average DPZ height estimates. The results demonstrate that the PINN-based state estimation can serve as a reliable soft sensor, enabling the real-time monitoring of this separation process from readily available flow-rate measurements.

A few limitations of the present study should be noted. First, the model assumes a band-shaped DPZ, which becomes inaccurate at higher flow rates where the DPZ develops a distinct wedge-shaped profile. Although a mapping between the average and maximum DPZ height was learned by a simple neural network, the PINN model does not fully represent the actual separator dynamics. Second, the experimental data contain substantial gaps caused by missing phase-boundary detections, particularly for QIR03, which limits the training data quality. Third, the experimental dataset includes no trajectories with conditions close to flooding. While such conditions are difficult to obtain experimentally for safety reasons, their absence restricts the ability of the model to detect or classify flooding behavior. Fourth, we acknowledge that filtering the immeasurable internal flows in the proposed state estimation framework could improve phase height estimation accuracy, however this is not possible with the current PINN architecture. Lastly, the experimental dataset comprises only four trajectories, with only one being used for training.

These limitations outline clear directions for future work. Extending the framework to handle non-uniform DPZ geometries would improve predictive capability under a wider range of operating conditions. Experimental data quality for phase heights may be enhanced through improved illumination, refinements to the phase-boundary detection framework, and the use of larger training datasets. Entrained droplets in the outlets relevant to quantifying incomplete separation can also be investigated. Finally, near-flooding and flooding data would allow more rigorous evaluation of the estimation performance under safety-relevant operating conditions.

\section*{Declaration of Competing Interest}
We have no conflict of interest.

\setlength{\LTleft}{0pt}   
\setlength{\LTright}{0pt}  

\section*{Nomenclature}

\subsection*{Abbreviations}
\begin{longtable}{ll}
DPZdense-packed zone \\
EKF       & extended Kalman Filter \\
IDW       & inverse Dirichlet weighting \\
IQR       & interquartile range \\
L-BFGS    & limited-memory Broyden–Fletcher–Goldfarb–Shanno algorithm \\
LHS       & Latin hypercube sampling \\
LLS       & liquid-liquid separator \\
MSE       & mean squared error \\  
NN        & neural network \\
ODE       & ordinary differential equation \\
PDE       & partial differential equation \\
PINN      & physics-informed neural network \\
RMSE      & root mean square error \\
SGD       & stochastic gradient descent \\
UKF       & unscented Kalman Filter \\
VNN       & vanilla neural network \\
\end{longtable}

\subsection*{Greek Symbols}
\begin{longtable}{ll}
$\gamma$        & interfacial tension \\
$\epsilon$    & holdup \\
$\bm{\theta}$   & learnable parameters \\
$\lambda$       & PINN loss term weight \\
$\eta$           & dynamic viscosity \\
$\rho$          & density \\
$\Delta\rho$    & density difference between organic and aqueous phase \\
$\sigma$        & standard deviation \\
$\tau$          & process time \\
\end{longtable}

\subsection*{Latin Symbols}
\begin{longtable}{ll}
$d$   & diameter \\
$\mathcal{D}$   & dataset \\
$e$        & error \\
$\bm f$        & differential equation right-hand side \\
$\mathbf{F}$    & Jacobian of state transition model \\
$\bm g$        & algebraic equation right-hand side \\
$h$        & height \\
$\bm h$    & measurement model \\
$\mathbf{H}$    & Jacobian of measurement model \\     
$\mathbf{I}$    & identity matrix \\  
$\mathbf{K}$    & Kalman gain matrix \\
$L$        & separator length \\
$n$        & number \\
$\mathbf{P}$    & state covariance matrix \\
$\mathrm{PINN}_{\bm{\theta}}$ & PINN functional representation \\
$Q$        & volume flow rate \\
$\mathbf{W}$    & process noise covariance matrix \\
$r$        & separator radius \\
$\mathbf{R}$    & measurement noise covariance matrix \\
$t$        & PINN time \\
$T$        & time duration \\
$\Vu$      & control variables \\
$\mathbf{v}$ & measurement noise \\
$\mathbf{w}$ & process noise \\
$\Vx$      & state variables \\
$\Vy$      & system outputs \\
$\Vz$      & immeasurable internal flows \\
\end{longtable}

\subsection*{Subscripts}
\begin{longtable}{ll}
$0$            & initial state \\
$\mathrm{32}$  & Sauter mean \\
$\mathrm{bot}$ & bottom outflow \\
$\mathrm{c}$            & coalescence \\
$\mathrm{data}$ & measurement data \\
$\mathrm{DP}$ & dense-packed zone \\
$\mathrm{ensemble}$ & ensemble \\
$\mathrm{exp}$ & experimental \\
$\mathrm{g}$ & algebraic equation \\
$\mathrm{HP}$ & heavy phase \\
$\mathrm{in}$ & inlet \\
$k$ & time step \\
$\mathrm{max}$ & maximum \\
$\mathrm{out}$ & output \\
$\mathrm{physics}$ & physics \\
$\mathrm{init}$ & initial condition \\
$\mathrm{s}$           & sedimentation \\
$\mathrm{scale}$       & scaling \\
$\mathrm{selfsimilar}$ & self-similarity \\
$\mathrm{sensor}$ & sensor \\
$\mathrm{sep}$ & separator \\
$\mathrm{sim}$ & simulation \\
$\mathrm{swarm}$ & swarm \\
$\bm{\theta}$   & learnable parameters \\
$\mathrm{top}$ & top outflow \\
$z$ & immeasurable internal flows \\ 
\end{longtable}

\section*{Acknowledgements}
\label{sec:acknowledgements}

This work was funded by the Deutsche Forschungsgemeinschaft (DFG, German Research Foundation) – 466656378 – within the Priority Programme ``SPP 2331:Machine Learning in Chemical Engineering''. This work was performed as part of the Helmholtz School for Data Science in Life, Earth and Energy (HDS-LEE). We acknowledge financial support by the Helmholtz Association of German Research Centers through program-oriented funding.

\section*{Declaration of Generative AI and AI-Assisted Technologies}  
During the preparation of this work, generative AI tools (ChatGPT, Claude) were used to assist with grammar, spelling, and style improvements in the writing. Generative AI coding tools (Claude) were used for post-processing tasks, specifically for plotting and visualization of the results. In all cases, the authors reviewed and edited the outputs, and they take full responsibility for the integrity and accuracy of the publication.  

\section*{Author contributions}

\begin{itemize}
    \item MV developed the PINN-based dynamic model, implemented the PINN model into the Extended Kalman Filter framework, produced and analyzed the results, and wrote the draft of all sections except the experimental setup (\cref{sec:settler}) and the separation related parts of the introduction (\cref{sec:intro}).
    \item SZ designed the experimental setup and conducted experimental campaigns for data acquisition, implemented and extended the mechanistic model of the liquid-liquid separator based on the literature \citep{Backi2018AFirst-Principles}, and wrote the draft of experimental setup (\cref{sec:settler}) and the separation related parts of the introduction (\cref{sec:intro}).
    \item MD conceptualized and supervised the whole work with the exception of the derivation of the mechanistic separator model, and provided help and guidance on the methodology.
    \item AlMi provided conceptual input on the data pre-processing, PINN and Extended Kalman Filter and provided further supervision.
    \item AdMh provided conceptual input and guidance on the state estimation framework (\cref{sec:EKF}) and phase height estimation results (\cref{sec:results_discussion}). 
    \item AJ provided conceptual input on the separator model and provided further supervision.
    \item All authors have reviewed and edited the manuscript.
\end{itemize}

\section*{CRediT authorship contribution statement}

\textbf{Mehmet Velioglu}: Conceptualization, Methodology, Software, Investigation, Writing - original draft, review \& editing, Visualization \noindent \textbf{Song Zhai}: Conceptualization, Methodology, Software, Investigation, Writing - original draft, review \& editing, Visualization. \noindent \textbf{Alexander Mitsos}: Conceptualization, Writing - review \& editing, Supervision. \noindent \textbf{Adel Mhamdi}: Methodology, Writing - review \& editing. \noindent \textbf{Andreas Jupke}: Conceptualization, Writing - review \& editing, Supervision, Funding acquisition.  \noindent \textbf{Manuel Dahmen}: Conceptualization, Methodology, Writing - review \& editing, Supervision, Funding acquisition.

\renewcommand{\refname}{Bibliography}  
\bibliography{ms.bib}
\bibliographystyle{elsarticle-harv}
\end{document}